%% file: pami.tex
\documentclass[10pt,journal,compsoc]{IEEEtran}
\ifCLASSOPTIONcompsoc
\usepackage[nocompress]{cite}
\else
\usepackage{cite}
\fi
\ifCLASSINFOpdf
\else
\fi

\usepackage{epsfig}
\usepackage{graphicx}
\usepackage{amsmath}
\usepackage{amssymb}
\usepackage{enumitem}

\usepackage[pagebackref=true,breaklinks=true,letterpaper=true,colorlinks,bookmarks=false,urlcolor=red,citecolor=blue]{hyperref}
\usepackage{gensymb,graphics,subfigure,inputenc}
\usepackage[linesnumbered,algo2e,boxed]{algorithm2e}
\usepackage{ragged2e} 
\graphicspath{{Figure/}}
\usepackage[figuresright]{rotating}
\usepackage{amsmath,amssymb,textcomp,subfigure,multirow,upgreek}
\usepackage{booktabs}
\usepackage{color,mdwlist}

\usepackage{makecell}
\usepackage{utfsym}
\usepackage{arydshln}
\usepackage{cleveref}
\usepackage{threeparttable,booktabs}

\usepackage{graphicx,fontawesome}
\graphicspath{{Figures/}}

\usepackage{caption}
\usepackage{enumitem}
\setitemize{noitemsep,topsep=0pt,parsep=0pt,partopsep=0pt}

\begin{document}
	%
	\title{Physical Adversarial Attack Meets Computer Vision: A Decade Survey}
	\author{Hui~Wei, Hao~Tang, Xuemei Jia, Zhixiang Wang, \\Hanxun Yu, Zhubo Li, Shin'ichi Satoh, Luc Van Gool, and Zheng Wang
		\IEEEcompsocitemizethanks{
            \IEEEcompsocthanksitem Hui Wei, Xuemei Jia, and Zheng Wang are with the School of Computer Science, National Engineering Research Center for Multimedia Software, Wuhan University, Wuhan, P.R.China.  \protect
            \IEEEcompsocthanksitem Hao Tang is with the National Key Laboratory for Multimedia Information Processing, School of Computer Science, Peking University, Beijing
            100871, P.R.China.  \protect
            \IEEEcompsocthanksitem Hanxun Yu is with the Colleage of Software and Technology, Zhejiang University, Hangzhou, P.R.China.  \protect
            \IEEEcompsocthanksitem Zhubo Li is with the School of Cyber Science and Engineering, Wuhan University, Wuhan, P.R.China.  \protect
            \IEEEcompsocthanksitem Zhixiang Wang and Shin'ichi Satoh are with the Digital Content and Media Sciences Research Division, National Institute of Informatics, Japan, and also with the Department of Information and Communication Engineering, Graduate School of Information Science and Technology, The University of Tokyo, Japan.  \protect
            \IEEEcompsocthanksitem Luc Van Gool is with the Computer Vision Lab of ETH Zurich, 8092 Zürich, Switzerland, and also with KU Leuven, 3000 Leuven, Belgium, and INSAIT, Sofia. \protect 
			\IEEEcompsocthanksitem Zheng Wang is the corresponding author. E-mail: wangzwhu@whu.edu.cn \protect
		}
	}
	
	%
	%
	
	\markboth{IEEE Transactions on Pattern Analysis and Machine Intelligence}%
	{Shell \MakeLowercase{\textit{et al.}}: Bare Demo of IEEEtran.cls for Computer Society Journals}
	%



	\IEEEtitleabstractindextext{%
		\input{0abstract}
		
		}

	\maketitle

	\IEEEdisplaynontitleabstractindextext

	%
	\IEEEpeerreviewmaketitle


	%
	%
	%
	%
	
	\input{1introduction}

    \input{2preliminaries}

    \input{3mediums}

    \input{4evaluation}

    \input{5tasks}

    \input{6discussion}

    \input{7conclusion}
    
	\small
	\bibliographystyle{IEEEtran}
	\bibliography{reference}

\end{document}

%% file: 0abstract.tex
\justify
\begin{abstract}
Despite the impressive achievements of Deep Neural Networks (DNNs) in computer vision, their vulnerability to adversarial attacks remains a critical concern. Extensive research has demonstrated that incorporating sophisticated perturbations into input images can lead to a catastrophic degradation in DNNs' performance. This perplexing phenomenon not only exists in the digital space but also in the physical world. Consequently, it becomes imperative to evaluate the security of DNNs-based systems to ensure their safe deployment in real-world scenarios, particularly in security-sensitive applications. To facilitate a profound understanding of this topic, this paper presents a comprehensive overview of physical adversarial attacks. Firstly, we distill four general steps for launching physical adversarial attacks. Building upon this foundation, we uncover the pervasive role of artifacts carrying adversarial perturbations in the physical world. These artifacts influence each step. To denote them, we introduce a new term: adversarial medium. Then, we take the first step to systematically evaluate the performance of physical adversarial attacks, taking the adversarial medium as a first attempt. Our proposed evaluation metric, \textit{hiPAA}, comprises six perspectives: \textit{Effectiveness}, \textit{Stealthiness}, \textit{Robustness}, \textit{Practicability}, \textit{Aesthetics}, and \textit{Economics}. We also provide comparative results across task categories, together with insightful observations and suggestions for future research directions.
\end{abstract}

%% file: 1introduction.tex
\IEEEraisesectionheading{\section{Introduction}
	\label{sec:introduction}}

\IEEEPARstart{D}{eep} Neural Networks (DNNs) have achieved impressive results in a variety of fields: from computer vision~\cite{voulodimos2018deep} to natural language processing~\cite{otter2020survey} to speech processing~\cite{nassif2019speech}, and it is increasingly empowering many aspects of modern society. Nonetheless, Szegedy \textit{et al.} \cite{SzegedyZSBEGF13} discovered in 2014 that adversarial samples can cause DNNs-based models to produce incorrect predictions, leading to a significant degradation in performance.
This is a groundbreaking work that exposes the vulnerability of DNNs, casting a shadow over their reliability and security.
Since then, researchers have conducted extensive explorations into adversarial samples, revealing their pervasive existence across a wide range of DNNs-based computer vision tasks~\cite{carlini2017towards,xiao2018generating,inkawhich2019feature,yuan2019adversarial,zhao2020towards,diao2021basar,cai2022zero,wei2022towards}.
They designed adversarial clothing~\cite{thys2019fooling} to evade person detectors, adversarial eyeglasses~\cite {sharif2016accessorize} to deceive face recognizers, etc. Increasingly, these methods use a class of techniques called adversarial attacks.

\begin{table}[t]
\caption{Comparative analysis of current surveys on physical adversarial attack in computer vision.}
\label{tab:comparison}
\centering
\tabcolsep=3pt
\resizebox{1.0\linewidth}{!}{
\begin{tabular}{lcccccc}
    \toprule
    {Survey} &  {\makecell[c]{Physical\\ Attack}} & {\makecell[c]{Adversarial\\ Medium}} & {Evaluation} &  {\makecell[c]{Number of\\ Methods}} & {\makecell[c]{Number of\\ Tasks}} & {Year}\\
    \midrule
    \cite{sun2018survey} & \usym{1F5F8} & \usym{2717}  &\usym{2717}& 5 & 3 & 2018\\
    \cite{wei2022physically} & \usym{1F5F8} & \usym{2717}  &\usym{2717}& 47 & 11 & 2022\\
    \cite{wang2022survey} & \usym{1F5F8} & \usym{2717}  &\usym{2717}& 69 & 7 & 2022\\
    \cite{nguyen2023physical} & \usym{1F5F8} & \usym{2717}  &\usym{2717}& 22 & 4 & 2023 \\
    Ours & \usym{1F5F8} & \usym{1F5F8}  &\usym{1F5F8}& {\color{black}86} & {\color{black}15} & {\color{black}2024}\\
    
    \bottomrule
\end{tabular}}
\vspace{-0.4cm}
\end{table}
 
Generally, adversarial attacks occur by adding 
perturbations to input data (e.g., image, video) and fooling the DNNs-based models in the inference stage.
Regarding the various domains, adversarial attacks can be categorized into two distinct classes: 
\textbf{(i) Digital Adversarial Attack}, 
which occurs in the digital space through the addition of subtle perturbations (e.g., style perturbations \cite{xu2021towards} and context-aware perturbations \cite{cai2022context}). 
\textbf{(ii) Physical Adversarial Attack}, 
which occurs in the real world using tangible artifacts that contain adversarial perturbations (e.g., adversarial patches \cite{brown2017adversarial,tan2021legitimate} and adversarial stickers \cite{zhang2018camou,komkov2021advhat}).
Compared to the former, the latter pose an augmented threat to social security, raising significant apprehensions, particularly in safety-critical domains like autonomous driving, video surveillance, and facial biometric systems.
To facilitate a profound understanding and provide in-depth insights into this topic, we present a comprehensive review of articles on physical adversarial attacks in computer vision tasks.

\begin{figure*}[t]
  \centering
  \includegraphics[width=0.99\linewidth]{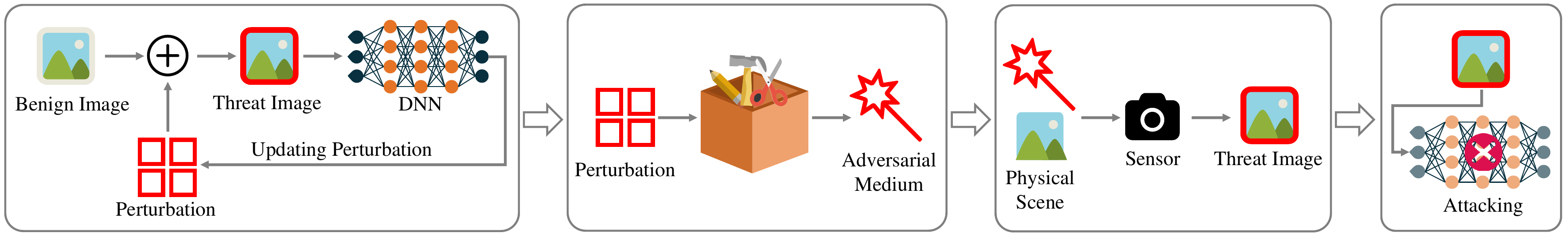}
  \caption{The flow of designing a physical adversarial attack, including four main steps: 1) \textit{Adversarial perturbation generation in the digital space}, 2) \textit{Adversarial medium manufacturing in the physical space}, 3) \textit{Threat image capturing}, and 4) \textit{Attacking}.}
  \label{figure:workflowPPA}
  \vspace{-0.4cm}
\end{figure*}

Though some existing surveys have also summarized the physical adversarial attack methods \cite{sun2018survey,wei2022physically,wang2022survey,nguyen2023physical}, 
they primarily focus on listing and categorization, ignoring the evaluation and comparison (see TABLE \ref{tab:comparison}).
A unified evaluation criterion is still absent.
This motivates us to take the first step to evaluate the performance of physical adversarial attacks systematically.

To this end, we outline the four general steps (as shown in Fig.~\ref{figure:workflowPPA}) required to build a physical adversarial attack:
\begin{enumerate}
    \item Step 1: \textit{Adversarial perturbation generation}. Generating perturbations in the digital domain based on given DNNs-based models, constrained by different attack forms and attack objectives.
    \item Step 2: \textit{Adversarial medium manufacturing}. Designing appropriate physical medium for carrying the perturbations in alignment with attack forms, and subsequently manufacturing them using suitable materials.
    \item Step 3: \textit{Threat image capturing}. Applying the adversarial medium in real-world scenarios to be captured by an imaging sensor, thereby generating threat images.
    \item Step 4: \textit{Attacking}. The captured threat images are fed to the DNNs-based model to initiate attacks.
\end{enumerate}

Note that we introduce the concept of an ``adversarial medium'' to denote the tangible artifact responsible for carrying the adversarial perturbation in the real world.
According to the four steps mentioned above, we discern the significant role of adversarial mediums in building a physical adversarial attack. 
They determine the form of perturbations (Step 1), impact manufacturing processes (Step 2), and hold relevance for real-world applications (Step 3).
Therefore, we embrace an approach centered on adversarial mediums to examine the existing methods, systematically quantifying and evaluating them in the following six perspectives: 
\textit{Effectiveness}, \textit{Stealthiness}, \textit{Robustness}, \textit{Practicability}, \textit{Aesthetics}, and \textit{Economics}.
Meanwhile, we introduce a comprehensive metric, the hexagonal indicator of Physical Adversarial Attack (\textit{hiPAA}), and provide comparative results across task categories, along with insightful observations and suggestions for future research directions.

Our contributions can be summarized as follows:
\begin{enumerate}
\item Through a comprehensive review of existing methodologies (see TABLEs~\ref{tab:stickerPatch},~\ref{tab:clothingImageLight},~\ref{tab:nicheMediums}), we abstract and summarize a general workflow for launching a physical adversarial attack, comprising four distinct steps (see Fig.~\ref{figure:workflowPPA}).
\item Leveraging this general workflow, we discover that tangible artifacts carrying adversarial perturbations exert substantial influence over attacks, prompting the introduction of the new concept of adversarial medium to represent them. 
\item As opposed to existing reviews \cite{sun2018survey,wei2022physically,wang2022survey,nguyen2023physical}, we take the first step to systematically evaluate the performance of physical adversarial attack, taking adversarial medium as a first attempt. Our proposed metric, \textit{hiPAA}, comprises six perspectives: \textit{Effectiveness}, \textit{Stealthiness}, \textit{Robustness}, \textit{Practicability}, \textit{Aesthetics}, and \textit{Economics}.
\item We conduct comprehensive comparisons of existing attack methods, and discuss limitations, challenges, and potential directions from the standpoint of real-world applications to facilitate future research.
\end{enumerate}

The remaining article is organized as follows. We first provide a brief introduction to the preliminaries in Sec. \ref{sec:preliminaries}, which cover essential topics and concepts for the proper understanding of this work. Sec.~\ref{sec:mediums} discusses the concept of adversarial mediums. Sec.~\ref{sec:evaluation} introduces the evaluation metric for conducting a comprehensive comparison. Then we present the recent advancements in physical adversarial attacks according to mainstream visual tasks and systematically evaluate them for each task in~Sec. \ref{sec:tasks}. Moreover, we provide discussions and future research opportunities in Sec.~\ref{sec:discussion}. Finally, we conclude this review in Sec.~\ref{sec:conclusion}. An overview of the survey scope is shown in Fig.~\ref{figure:content}. We also provide a regularly updated project page on \url{https://github.com/weihui1308/PAA}.

%% file: 2preliminaries.tex
\section{Preliminaries}
	\label{sec:preliminaries}

In this section, we provide a concise introduction to problem formulation, key topics, and concepts, aiming to enhance comprehension of our work. 

\begin{figure*}[!t]
  \centering
  \includegraphics[width=0.98\linewidth]{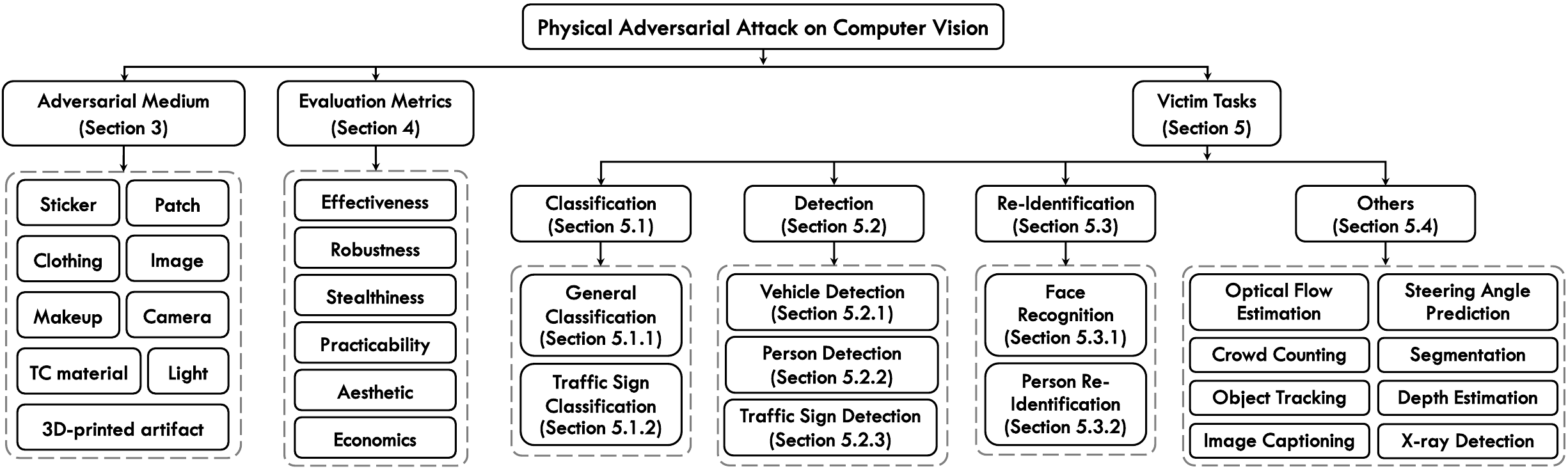}
  \caption{{\color{black}A general overview of the scope in our survey.}}
  \label{figure:content}
  \vspace{-0.4cm}
\end{figure*}

\subsection{Computer Vision}
Computer Vision (CV) aims to enable machines to perceive, observe, and understand the physical world like human eyes.
An important milestone was reached when Krizhevsky \textit{et al}. \cite{russakovsky2015imagenet} proposed AlexNet, which secured victory in the ILSVRC, thus promoting the application of DNNs to address a wide variety of tasks.
Up to the present, DNNs-based models have achieved impressive and competitive performances in CV tasks, including  classification~\cite{liu2021swin,yu2022coca}, 
segmentation~\cite{he2017mask,badrinarayanan2017segnet}, 
detection~\cite{redmon2018yolov3,zhang2022dino}, 
and re-identification~\cite{ye2018visible,wang2018cascaded,meng2021magface}.
This survey concentrates on three prominent tasks: classification, detection, and re-identification, which are extensively utilized and encompass numerous attack scenarios.
Given a DNNs-based model $f{:}X{\to} {Y}$ with pre-trained weights $\theta$, for the sake of conciseness, we formulate the DNNs-based CV models as follows:
\begin{equation}
  {\color{black}\hat{y}=f_\theta(x),\qquad x\in X,y\in Y,}
\end{equation}
where for any input data $x\in X$, {\color{black}the well-trained model $f_\theta( \cdot )$ is able to predict a $\hat{y}$ that closely approximates the corresponding ground truth $y\in Y$.}

\subsection{Adversarial Attacks}
\subsubsection{Problem Formulation}
{\color{black}Adversarial attacks involve introducing perturbations to input data, causing the model $f_\theta(\cdot)$ to produce incorrect predictions $y'$.} Note that the attacker's modification is limited to the input data. Exactly, the sample $x$, following the addition of perturbations $\delta$, is denoted as the adversarial sample $x'$. Mathematically, the joint representation is expressed as
\begin{equation}
\begin{cases}
x'=x + \delta \\
{\color{black}y'=f_\theta(x')}
\end{cases}
s.t. \;\; y'\ne \hat{y},
\end{equation}
where perturbations $\delta$ are typically constrained by factors such as intensity, size, and the adversarial medium.

\subsubsection{Distinguishing Adversarial Attacks from Backdoor Attacks and Poisoning Attacks.}
\begin{figure*}[t]
  \centering
  \includegraphics[width=0.87\linewidth]{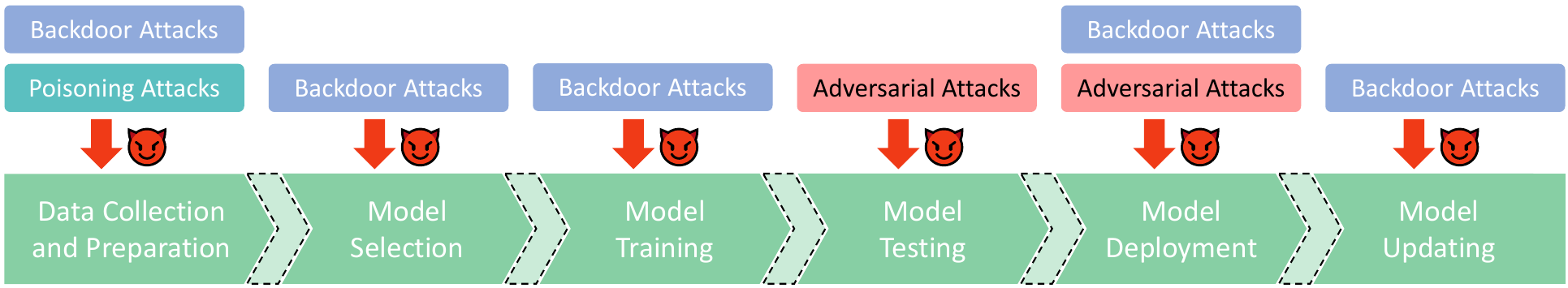}
  \caption{An overview of the lifecycle in which the three types of attacks occur. Adversarial attacks occur only during the model deployment phase, without modifying the model and training data. Compared to backdoor attacks and poisoning attacks, adversarial attacks have weaker assumptions, focusing on the vulnerability of the model itself.}
  \label{figure:lifecycle}
  \vspace{-0.4cm}
\end{figure*}

Apart from adversarial attacks, there are two other widely used attack types: backdoor (a.k.a. trojan) attacks~\cite{gu2017badnets,liu2017trojaning,wenger2021backdoor,qi2022towards} and poisoning attacks~\cite{barreno2006can,shafahi2018poison,zhang2020online}.
While these three attack categories aim to mislead the model into incorrect predictions, their implementation methods differ fundamentally. We elucidate these distinctions by introducing the DNNs-based model lifecycle, depicted in Fig. \ref{figure:lifecycle}, which comprises six phases: data collection and preparation, model selection, model training, model testing, model deployment, and model updating.

\noindent{\textbf{Poisoning Attack.}}
DNN-based models require substantial data to achieve high performance. Developers often use publicly available datasets or scrape data from online sources, creating opportunities for poisoning attacks. These attacks involve injecting malicious or misleading samples into the training set, which can diminish learning efficiency and, in some cases, hinder convergence, ultimately disrupting the learning process~\cite{oprea2022poisoning}. Poisoning attacks typically occur during the data collection and preparation phase.

\noindent{\textbf{Backdoor Attack.}}
Backdoor attacks involve injecting a hidden pattern or trigger into a DNNs-based model, causing incorrect behavior or erroneous predictions when the model encounters input data containing the trigger. Attackers achieve this by contaminating training data, altering model weights, or modifying the model architecture~\cite{gao2020backdoor}. These attacks can occur at multiple stages in the model lifecycle, using methods such as code poisoning~\cite{bagdasaryan2021blind}, data poisoning~\cite{gu2017badnets}, and controlling the training process~\cite{doan2021lira}. Backdoor attacks should remain inconspicuous to users and should not disrupt the normal functioning of DNNs.

\noindent{\textbf{Adversarial Attacks.}}
Compared to the two attack types mentioned above, adversarial attacks have the weakest underlying assumption: they solely modify input data, without any other alterations. This characteristic facilitates attackers in conducting practical applications more easily. Adversarial attacks occur during both the model testing and model deployment phases.

\subsection{The Taxonomy of Adversarial Attacks}
Adversarial attacks can generally be categorized into three types based on adversarial knowledge:

\noindent{\textbf{White-Box Attacks.}}
In a white-box attack, adversaries have full access to data and model details, including network architecture, parameters, and weights. They exploit this information to assess vulnerabilities and adapt their strategies, making this attack method relatively straightforward to execute.

\noindent{\textbf{Black-Box Attacks.}}
{\color{black}The attackers lack access to the target model's structure and parameters and can only interact with the model to obtain predictions for specific samples.}

\noindent{\textbf{Gray-Box Attacks.}} 
{\color{black}In a gray-box setting, the attack lies between black-box and white-box attacks, where the attacker has partial knowledge of the model, such as the model's output probability distribution or model structure.}

Considering the adversarial goals, adversarial attacks can be divided into two categories:

\noindent{\textbf{Targeted Attacks.}}
Targeted attacks aim to mislead DNN-based models to specified labels, e.g., misidentifying a cat image as that of a dog. The specified labels make targeted attacks more challenging to achieve with high success rates. 

\noindent{\textbf{Untargeted Attacks.}}
The untargeted attacks mislead DNNs-based models to any wrong label. 
Designing an effective untargeted attack only requires making the model's predictions incorrect, without necessarily concerning what the incorrect predictions are.

\subsection{Introduction to Physical Adversarial Attacks}
{\color{black}Physical adversarial attacks refer to where attackers deploy tangible perturbations in physical space to mislead DNNs-based models into providing inaccurate predictions.}
As shown in Fig.~\ref{figure:workflowPPA}, we summarize the general process of physical adversarial attack as four steps: 1) \textit{Adversarial perturbation generation}, 2) \textit{Adversarial medium manufacturing}, 3) \textit{Threat image capturing}, and 4) \textit{Attacking}.

Unlike digital adversarial attacks, which typically involve imperceptible perturbations to the human eye, 
as demonstrated in the One Pixel Attack~\cite{su2019one},
physical adversarial attacks demand more intense perturbations. 
This level of intensity is crucial for the perturbation to be detectable by sensors in real-world scenarios.
Meanwhile, physical adversarial attacks face additional challenges due to various physical constraints (e.g., spatial deformation) and environmental dynamics (e.g., lighting). 

It is worth noting that backdoor attacks can also be executed in the physical world~\cite{02339Ma48550,wei2023moire}. For instance, Qi \textit{et al.}~\cite{qi2022towards} proposed a physically realizable backdoor attack algorithm against image classification tasks. However, it's important to clarify that this category is not within the scope of our discussion, as our primary focus is on physical adversarial attacks.

\begin{table*}
    \centering
    \caption{
        {{\color{black}Chronological overview of the typical physical adversarial attach methods using different adversarial mediums. We provide detailed definitions for each adversarial medium in this survey. Colors indicate the category of adversarial mediums used. $\ast$ denotes milestone papers.} }
    }
    \label{tab:mediumdefinitions}
    \begin{tabular}{clp{8.0cm}}
        \toprule
        Chronological overview & Adversarial medium   & Definition in this survey \\
        \midrule
        \multirow{9}{*}{\begin{minipage}[b]{0.55\columnwidth}
		\centering
		\raisebox{-.7\height}{\includegraphics[width=\linewidth]{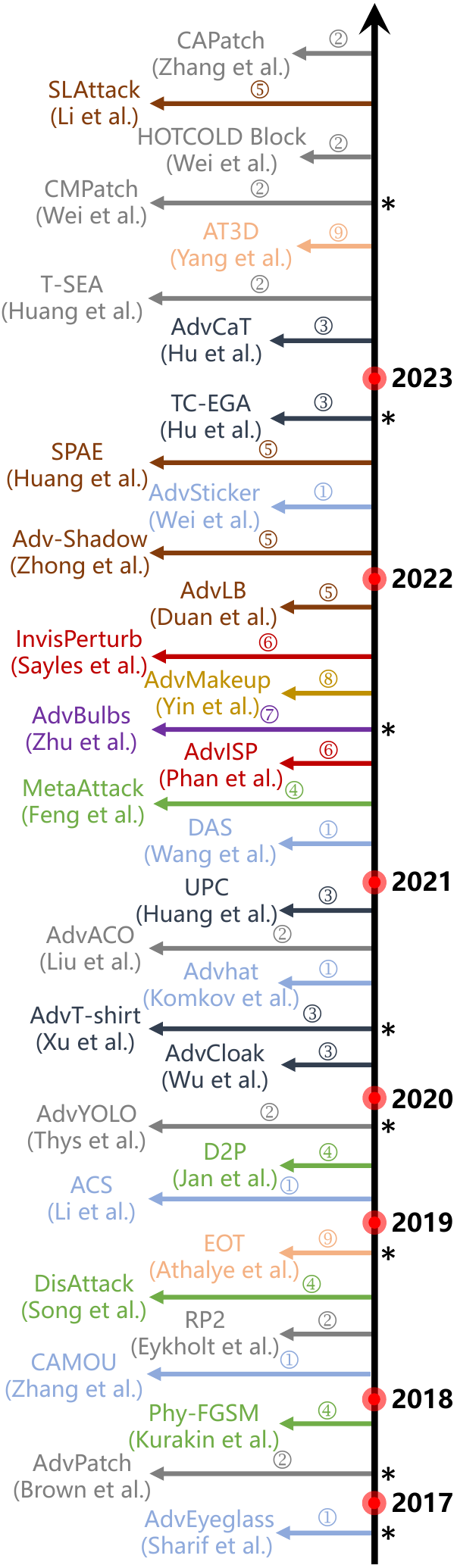}}
	\end{minipage}} & 
        \begin{minipage}[b]{0.24\columnwidth}
		\centering
		\raisebox{-.7\height}{\includegraphics[width=\linewidth]{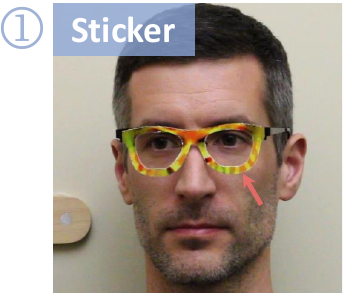}}
	\end{minipage}  & Sticker is \textbf{a piece of paper or plastic}, with adversarial perturbation patterns on one side, and needs to be \textbf{affixed} to a host object. There are no size restrictions, and it can cover the entire surface if necessary. \\
        \specialrule{0em}{1pt}{1pt}
        \cline{2-3}
        \specialrule{0em}{1pt}{1pt}
        & \begin{minipage}[b]{0.24\columnwidth}
		\centering
		\raisebox{-.7\height}{\includegraphics[width=\linewidth]{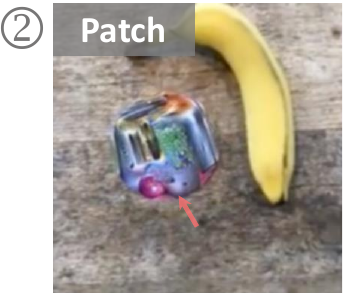}}
	\end{minipage}   & Patch is a material with \textbf{limited area} that carries perturbations, which can be in the form of \textbf{paper or cardboard}. Its area is \textbf{smaller} than the object to which it is applied.\\
        \specialrule{0em}{1pt}{1pt}
        \cline{2-3}
        \specialrule{0em}{1pt}{1pt}
        & \begin{minipage}[b]{0.24\columnwidth}
		\centering
		\raisebox{-.7\height}{\includegraphics[width=\linewidth]{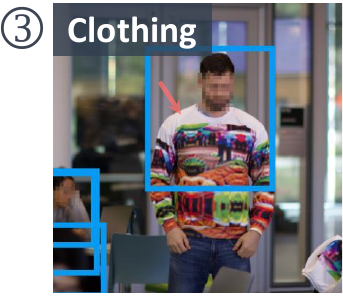}}
	\end{minipage}  &  Clothing as an adversarial medium refers to the attacker imprinting or crafting perturbed textures onto \textbf{garments}, constituting \textbf{a wearable design}. Examples include t-shirts, capes, or pants.\\
        \specialrule{0em}{1pt}{1pt}
        \cline{2-3}
        \specialrule{0em}{1pt}{1pt}
        & \begin{minipage}[b]{0.24\columnwidth}
		\centering
		\raisebox{-.7\height}{\includegraphics[width=\linewidth]{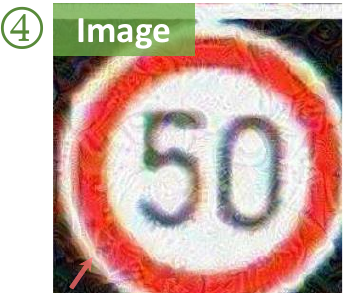}}
	\end{minipage}  & Image as an adversarial medium entails the attacker incorporating adversarial perturbations \textbf{across the entire image}. In contrast, stickers and patches are applied exclusively to the target object or a limited area.\\
        \specialrule{0em}{1pt}{1pt}
        \cline{2-3}
        \specialrule{0em}{1pt}{1pt}
        & \begin{minipage}[b]{0.24\columnwidth}
		\centering
		\raisebox{-.7\height}{\includegraphics[width=\linewidth]{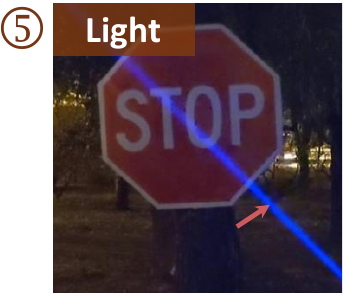}}
	\end{minipage}  & Light as the adversarial medium refers to the attacker using \textbf{light} to construct adversarial perturbations in the physical space and then \textbf{projecting} them onto the target object. \\
        \specialrule{0em}{1pt}{1pt}
        \cline{2-3}
        \specialrule{0em}{1pt}{1pt}
        & \begin{minipage}[b]{0.24\columnwidth}
		\centering
		\raisebox{-.85\height}{\includegraphics[width=\linewidth]{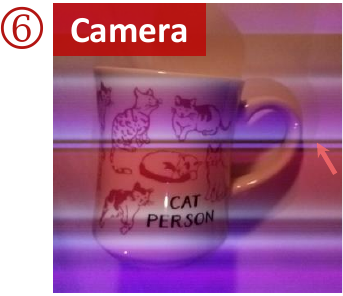}}
	\end{minipage}  & Camera as an adversarial medium refers to attackers exploiting the inherent \textbf{imaging characteristics} or \textbf{parameter settings} of the camera to construct adversarial perturbations. This approach involves introducing adversarial perturbations during the capture of physical scenes, thereby affecting downstream tasks.\\
        \specialrule{0em}{1pt}{1pt}
        \cline{2-3}
        \specialrule{0em}{1pt}{1pt}
        & \begin{minipage}[b]{0.24\columnwidth}
		\centering
		\raisebox{-.75\height}{\includegraphics[width=\linewidth]{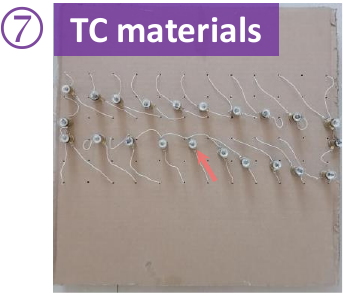}}
	\end{minipage} &  TC materials as an adversarial medium, involve attackers leveraging specialized physical materials to influence the imaging of thermal infrared cameras. This primarily affects \textbf{the distribution of thermal radiation}, enabling the construction of adversarial textures.\\
        \specialrule{0em}{1pt}{1pt}
        \cline{2-3}
        \specialrule{0em}{1pt}{1pt}
        & \begin{minipage}[b]{0.24\columnwidth}
		\centering
		\raisebox{-.7\height}{\includegraphics[width=\linewidth]{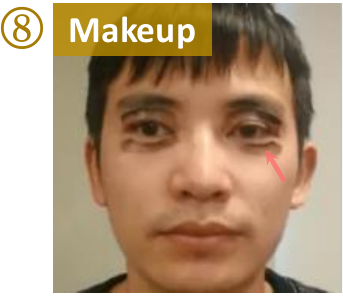}}
	\end{minipage}  &  Makeup as an adversarial medium refers to attackers using \textbf{cosmetic techniques} to alter facial appearance by applying pre-generated adversarial perturbations to the face.\\
        \specialrule{0em}{1pt}{1pt} 
        \cline{2-3}
        \specialrule{0em}{1pt}{1pt}
        & \begin{minipage}[b]{0.24\columnwidth}
		\centering
		\raisebox{-.75\height}{\includegraphics[width=\linewidth]{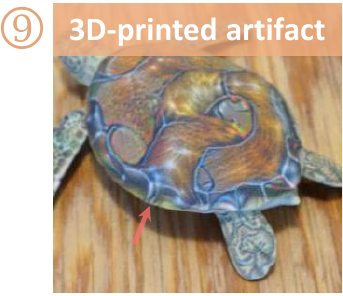}}
	\end{minipage}  & 3D-printed artifact as an adversarial medium refers to attackers employing \textbf{3D printing technology} to fabricate adversarial objects with a 3D spatial structure. Such adversarial objects are typically designed to execute real-world attacks from multiple angles.\\
        \bottomrule
    \end{tabular} 
        \vspace{-0.4cm}
\end{table*}

%% file: 3mediums.tex
\section{Adversarial Mediums}
	\label{sec:mediums}
We observe a commonality across all physical adversarial attack methods: the necessity of a physical entity to carry the specially designed perturbations.
Therefore, we introduce the novel concept of adversarial medium for the first time to represent these entities.
The \emph{medium} is commonly used in physics. When a substance exists inside another substance, the latter is the medium of the former~\cite{van1985observation}. 
In most cases, properties such as form, density, and shape of the former would influence the properties of the latter. 
Analogously, for launching attacks in the physical space, the adversarial perturbation must have a carrier, i.e., the adversarial perturbation exists inside the carrier. Meanwhile, the carrier has an effect on the perturbation. Thus we define the carrier as the adversarial medium. 

An adversarial medium is indispensable for physical adversarial attacks. 
In TABLEs~\ref{tab:stickerPatch},~\ref{tab:clothingImageLight},~\ref{tab:nicheMediums}, we have compiled a comprehensive list of all the physical adversarial attack methods examined in this paper, organized by medium and chronological order.
Additionally, 
{\color{black}TABLE~\ref{tab:mediumdefinitions} enumerates nine adversarial mediums, which, to our knowledge, encompass all current types. We provide specific definitions for each adversarial medium in this review. Here, we specifically clarify the similarities and differences between stickers and patches, as they can be easily confused. Specifically, they share commonalities, such as 1) the general use of materials like paper or plastic, 2) application to the host object, and 3) limited area coverage. The distinctions lie in 1) the method of application, with stickers emphasizing attachment to the host object, while patches can involve sticking, hanging, or embedding, and 2) the relative size of the covered area, as stickers can cover the entire host object, whereas patches only occupy a small portion.
}

\begin{table*}
\setlength\tabcolsep{5pt}
  \caption{Physical adversarial attack methods that use \textbf{stickers} and \textbf{patches} as adversarial mediums. We list them by the adversarial medium and time order.}
  \label{tab:stickerPatch}
  \centering
\begin{tabular}{cllllccc}
    \toprule
    {\multirow{2}{*}{\makecell[c]{Adversarial\\ Medium}}}  & \multicolumn{3}{c}{{\makecell[c]{Description}}} &  \multirow{2}{*}{Method}  & \multirow{2}{*}{Victim Task} & \multirow{2}{*}{Venue}  & \multirow{2}{*}{Year} \\
    \cmidrule(lr){2-4}
    & Manufacture & Instrument & \makecell[c]{Attack Type} & &&& \\
    \midrule
    
\multirow{16}{*}{Sticker} & print, paste & eyeglasses & impersonation, dodging & AdvEyeglass~\cite{sharif2016accessorize}& Face Recognition & ACM CCS & 2016 \\

& cover & car body & hiding & CAMOU~\cite{zhang2018camou} & Vehicle Detection & ICLR & 2018 \\

& display & screen & hiding &  InvisibleCloak~\cite{yang2018building} & Person Detection & UEMCON & 2018 \\

& paste & camera lens & misclassification & ACS~\cite{li2019adversarial} & General Classification & PMLR & 2019 \\

& print, paste & eyeglasses & impersonation, dodging &  AdvEyeglass+ \cite{sharif2019general}  & Face Recognition   & TOPS & 2019 \\

& print, affix & hat & impersonation & Advhat~\cite{komkov2021advhat} & Face Recognition  & ICPR & 2020 \\

& print, paste & eyeglasses & misrecognition &  CLBAAttack \cite{singh2021brightness}  & Face Recognition   & BIOSIG & 2021   \\

& affix & car body & hiding & DAS~\cite{wang2021dual}   & \makecell[c]{Vehicle Detection}  & CVPR & 2021 \\

& affix & road marking & misdirection & AdvMarkings~\cite{272270}   & Lane Detection    & USENIX & 2021 \\

& full cover & car body & hiding & FCA~\cite{wang2022fca}   & Vehicle Detection & AAAI & 2022 \\

& full cover & car body & hiding & DTA~\cite{suryanto2022dta}  & Vehicle Detection & CVPR & 2022 \\

& imprint & face mask & dodging & AdvMask~\cite{zolfi2021adversarial} & Face Recognition & ECML PKDD & 2022 \\

& print, paste & face & impersonation & AdvSticker~\cite{wei2022adversarial}    & Face Recognition & TPAMI & 2022 \\

& cover & car body & hiding & CAC~\cite{duan2022learning}   & Vehicle Detection & IJCAI & 2022 \\

& {\color{black}print, paste} & {\color{black}face} & {\color{black}impersonation} & {\color{black}DOPatch~\cite{wei2023distributional}} & {\color{black}Face Recognition} & {\color{black}arXiv} & {\color{black}2023} \\

& {\color{black}print, paste} & {\color{black}car body} & {\color{black}false estimation} & {\color{black}$3D^2$Fool~\cite{zheng2024physical}} & {\color{black}Depth Estimation} & {\color{black}CVPR} & {\color{black}2024} \\

\specialrule{0em}{1pt}{1pt}
\hdashline
\specialrule{0em}{1pt}{1pt}

\multirow{28}{*}{Patch} & print, put & image patch & misclassification & AdvPatch~\cite{brown2017adversarial} & General Classification  & NIPS & 2017 \\

& print, paste & traffic sign & misclassification & RP$_2$~\cite{eykholt2018robust}   & Sign Classification    & CVPR & 2018 \\


& print, paste & traffic sign & misdetection & NestedAE~\cite{zhao2019seeing}    & Sign Detection    & CCS & 2019 \\

& print, paste & traffic sign & misclassification & PS-GAN~\cite{liu2019perceptual}    & Sign Classification    & AAAI & 2019 \\

& display & screen & lose track & PAT~\cite{wiyatno2019physical} & Object Tracking & ICCV & 2019 \\

& print & image patch & hiding & AdvYOLO~\cite{thys2019fooling} & Person Detection  & CVPRW & 2019 \\

& print, paste & image patch & mismatching & AdvPattern~\cite{wang2019advpattern}   & Person Re-ID      & ICCV & 2019 \\


& imprint & image patch & false estimation  & FlowAttack~\cite{ranjan2019attacking}   & Flow Estimation   & ICCV & 2019 \\

& print, paste & image patch & misclassification  & AdvACO~\cite{liu2020bias}   & General Classification    & ECCV & 2020 \\

& paste & camera lens & misdetection  & TransPatch~\cite{zolfi2021translucent}  & Sign Detection    & CVPR & 2021 \\

& print, paste & image patch & lose track  & MTD~\cite{ding2021towards}  & Object Tracking    & AAAI & 2021 \\

& print, paste & face & impersonation  & TAP~\cite{xiao2021improving}  & Face Recognition & CVPR & 2021 \\

& print, paste & image patch & misclassification & AdvACO+~\cite{wang2021universal}  & General Classification  & TIP & 2021 \\

& display & screen & misdetection & AITP~\cite{sava2022assessing}  & Sign Detection    & ACM AISec & 2022 \\

& print, paste & image patch & misclassification & CPAttack~\cite{casper2022robust}  & General Classification    & NIPS & 2022 \\

& print, paste & image patch & misclassification & TnTAttack~\cite{doan2022tnt}  & General Classification    & TIFS & 2022 \\

& print, paste & image patch & false estimation & OAP~\cite{cheng2022physical} & Depth Estimation  & ECCV & 2022 \\

& print & image patch & false segmentation & RWAEs~\cite{nesti2022evaluating}    & Segmentation      & WACV & 2022 \\

& print, paste & image patch & false estimation & PAP~\cite{liu2021harnessing}     & Crowd Counting    & ACM CCS & 2022 \\

& print, put & image patch & misclassification & DAPatch~\cite{chen2022shape} & General Classification    & ECCV & 2022 \\

& print, paste & face & impersonation & SOPP~\cite{wei2022simultaneously} & Face Recognition    & TPAMI & 2022 \\

& print, paste & car & hiding & AerialAttack~\cite{du2022physical}  & Vehicle Detection    & WACV & 2022 \\

& paste & aerogel patch & hiding & AdvInfrared~\cite{Wei_2023_CVPR} & Person Detection    & CVPR & 2023 \\

& display & screen & hiding & T-SEA~\cite{huang2022t} & Person Detection    & CVPR & 2023 \\

& paste & aerogel patch & hiding & CMPatch~\cite{wei2023unified}& Person Detection    & ICCV & 2023 \\

& print & image patch & misstatement& CAPatch~\cite{zhang2023capatch} & Image Captioning & USENIX & 2023 \\

& \color{black}{paste} & \color{black}{aerogel patch} & \color{black}{hiding} & \color{black}{IAPatch~\cite{wei2023infrared}} & \color{black}{Person Detection}    & \color{black}{IJCV} & \color{black}{2023} \\

& \color{black}{print, paste} & \color{black}{image patch} & \color{black}{mis\{det., cla.\}} & \color{black}{TPatch}~\cite{zhu2023tpatch} & \color{black}{Sign Det. \& Cla.} & \color{black}{USENIX} & \color{black}{2023} \\

    \bottomrule
\end{tabular}
  \vspace{-0.4cm}
\end{table*}

\begin{table*}
\setlength\tabcolsep{4.2pt}
  \caption{Physical adversarial attack methods that use \textbf{clothing}, \textbf{Images}, and \textbf{lights} as adversarial mediums. We list them by the adversarial medium and time order.}
  \label{tab:clothingImageLight}
  \centering
\begin{tabular}{cllllccc}
    \toprule
    {\multirow{2}{*}{\makecell[c]{Adversarial\\ Medium}}}  & \multicolumn{3}{c}{{\makecell[c]{Description}}} &  \multirow{2}{*}{Method}  & \multirow{2}{*}{Victim Task} & \multirow{2}{*}{Venue}  & \multirow{2}{*}{Year} \\
    \cmidrule(lr){2-4}
    & Manufacture & Instrument & \makecell[c]{Attack Type} & &&& \\
    \midrule
    
\multirow{9}{*}{Clothing} & print & T-shirt & hiding  & AdvT-shirt~\cite{xu2020adversarial} & Person Detection & ECCV & 2020 \\

& print & pants, sweaters, mask  & misdetection  & UPC~\cite{huang2020universal}  & Person Detection  & CVPR &2020 \\

& print & sweatshirt  & hiding  & AdvCloak~\cite{wu2020making}     & Person Detection  & ECCV &2020 \\

& print & sweatshirt  & hiding  & NAP~\cite{hu2021naturalistic}   & Person Detection  & ICCV &2021 \\

& print & T-shirt  & hiding  & LAP~\cite{tan2021legitimate}       & Person Detection  & ACM MM &2021 \\

& print & dresses, T-shirts, skirts  & hiding  & TC-EGA~\cite{hu2022adversarial}         & Person Detection  & CVPR &2022 \\

& crop & aerogel material  & hiding & InvisClothing~\cite{zhu2022infrared}    & Person Detection  & CVPR &2022 \\

& \color{black}{rendering} & \color{black}{full coverage}  & \color{black}{hiding} & \color{black}{DAC}~\cite{sun2023differential}    & \color{black}{Person Detection}  & \color{black}{NN} & \color{black}{2023} \\

& print & pants, sweatshirt  & hiding & AdvCaT~\cite{Hu_2023_CVPR}    & Person Detection  & CVPR & 2023 \\

\specialrule{0em}{1pt}{1pt}
\hdashline
\specialrule{0em}{1pt}{1pt}

\multirow{11}{*}{Image} & print & picture & misclassification & Phy-FGSM~\cite{kurakin2017adversarial}    & General Classification    & ICLR &2017 \\ 

& print & picture & misdetection & ShapeShifter~\cite{chen2018shapeshifter}& Sign Detection  & ECML PKDD &2018 \\

& print & picture & misdetection & RP$_2$+~\cite{220580Song}   & Sign Detection    & USENIX &2018 \\

& print & picture & misclassification & D2P~\cite{jan2019connecting}  & General Classification    & AAAI &2019 \\

& print & picture & misclassification & ABBA~\cite{guo2020watch}             & General Classification    & NIPS &2020 \\

& print & picture & false prediction & PhysGAN~\cite{kong2020physgan}          & Steering Angle Prediction   & CVPR &2020 \\

& print & picture & misclassification & AdvCam~\cite{duan2020adversarial}       & General Classification    & CVPR &2020 \\

& print & picture & hiding & LPAttack~\cite{yang2020beyond} & Sign Detection & AAAI & 2020 \\

& print & picture & misclassification & MetaAttack~\cite{feng2021meta}          & General Classification    & ICCV &2021 \\

& print & picture & misclassification & ViewFool~\cite{dong2022viewfool}          &  General Classification    & NIPS &2022 \\

& print & picture & misclassification  & Meta-GAN \cite{feng2023robust} & General Classification & TIFS & 2023 \\

\specialrule{0em}{1pt}{1pt}
\hdashline
\specialrule{0em}{1pt}{1pt}

\multirow{12}{*}{Light} &  project & projector & misclassification & PTAttack~\cite{nichols2018projecting}  & General Classification    & AAAI & 2018 \\

&  project & projector & misclassification & Poster~\cite{man2019poster} & General Classification & IEEE S\&P & 2019 \\

&  project & projector & impersonation & ALPA~\cite{nguyen2020adversarial}       & Face Recognition  & CVPR & 2020 \\

&  project & projector & hiding & SLAP~\cite{272218}    & Sign Detection    & USENIX & 2021 \\

&  project & projector & misclassification & OPAD~\cite{gnanasambandam2021optical}   & General Classification    & ICCV & 2021 \\

&  project & laser pointer & misclassification & AdvLB~\cite{duan2021adversarial}        & General Classification    & CVPR & 2021 \\

&  project & shadow & misclassification & Adv-Shadow~\cite{zhong2022shadows}      & Sign Classification    & CVPR & 2022 \\

&  project & projector & misclassification & SPAA~\cite{huang2022spaa}      & General Classification    & IEEE VR & 2022 \\

&  project & projector & hiding & AdvLight~\cite{10095895}   & Vehicle Detection    & ICASSP & 2023 \\

&  project & laser pointer & hiding & AdvLS~\cite{hu2023adversarial} & Sign Detection & PMLR & 2023 \\

&  project & projector & impersonation & SLAttack~\cite{Li_2023_CVPR}   & Face Recognition  & CVPR & 2023 \\

&  \color{black}{reflect} & \color{black}{mirror} & \color{black}{misclassification} & \color{black}{RFLA}~\cite{wang2023rfla}   & \color{black}{Sign Classification}  & \color{black}{ICCV} & \color{black}{2023} \\

    \bottomrule
\end{tabular}
  \vspace{-0.2cm}
\end{table*}

\begin{table*}
\setlength\tabcolsep{4.3pt}
  \caption{Physical adversarial attack methods that use niche adversarial mediums. We list them by the adversarial medium and time order.}
  \label{tab:nicheMediums}
  \centering
\begin{tabular}{cllllccc}
    \toprule
    {\multirow{2}{*}{\makecell[c]{Adversarial\\ Medium}}}  & \multicolumn{3}{c}{{\makecell[c]{Description}}} &  \multirow{2}{*}{Method}  & \multirow{2}{*}{Victim Task} & \multirow{2}{*}{Venue}  & \multirow{2}{*}{Year} \\
    \cmidrule(lr){2-4}
    & Manufacture & Instrument & \makecell[c]{Attack Type} & &&& \\
    \midrule

\multirow{2}{*}{Camera}     & capture  & rolling shutter effect & misclassification  & InvisPerturb~\cite{sayles2021invisible} & General Classification & CVPR & 2021 \\

& capture  & camera ISP & misclassification & AdvISP~\cite{phan2021adversarial}  & General Classification  & CVPR & 2021 \\

\specialrule{0em}{1pt}{1pt}
\hdashline
\specialrule{0em}{1pt}{1pt}

\multirow{2}{*}{TC material} & heating & cardboard & hiding &  AdvBulbs~\cite{zhu2021fooling} & Person Detection & AAAI & 2021 \\

& paste & warm/cool paste & hiding & HOTCOLD Block~\cite{wei2023hotcold} & Person Detection & AAAI & 2023 \\

\specialrule{0em}{1pt}{1pt}
\hdashline
\specialrule{0em}{1pt}{1pt}

\multirow{1}{*}{Makeup} & cosmetics & face & impersonation & AdvMakeup~\cite{yin2021adv} & Face Recognition & IJCAI & 2021 \\  

\specialrule{0em}{1pt}{1pt}
\hdashline
\specialrule{0em}{1pt}{1pt}

\multirow{5}{*}{\makecell[c]{3D-print\\ artifact}} & 3D-print & tangible turtle & misclassification & EOT~\cite{athalye2018synthesizing} & General Classification & PMLR & 2018 \\

& rendering & renderer & misclassification & 3DAttack~\cite{zeng2019adversarial}  & General Classification  & CVPR & 2019 \\

& 3D-print & tangible mesh & impersonation & AT3D~\cite{Yang_2023_CVPR} & Face Recognition & CVPR & 2023 \\

& \color{black}{3D-print} & \color{black}{adversarial metal} & \color{black}{hiding} & \color{black}{X-Adv}~\cite{liu2023x} & \color{black}{X-ray Item Detection} & \color{black}{USENIX} & \color{black}{2023} \\

& \color{black}{3D-print} & \color{black}{tangible object} & \color{black}{mis\{cla., det., cap.\}} & \color{black}{TT3D}~\cite{huang2024towards} & \color{black}{Cla., Det., Cap.} & \color{black}{CVPR} & \color{black}{2024} \\
    \bottomrule
\end{tabular}
  \vspace{-0.4cm}
\end{table*}

%% file: 4evaluation.tex
\section{Evaluation Metric}
	\label{sec:evaluation}
Despite the widespread attention given to physical adversarial attacks, their performance has been subject to inconsistent and case-by-case evaluations.
This phenomenon primarily stems from two factors:
\begin{enumerate}
    \item \textit{Challenges in Replication}: The creation of adversarial mediums, these physical entities, often incur high production costs and face various inherent factors during the manufacturing process. Factors such as the variability in printer models used for pattern printing and the difficulties in standardizing aspects like clothing style and color can influence the process. Furthermore, maintaining consistent real-world experimental conditions proves to be a daunting task. These challenges collectively hinder the replication of experiments conducted by others, making comparisons arduous.
    \item \textit{Varied Objectives}: Physical adversarial attack methods pursue diverse dimensions of enhancement. Some methods prioritize increasing attack effectiveness~\cite{brown2017adversarial,thys2019fooling,wei2023unified}, while others place emphasis on improving attack stealthiness~\cite{liu2019perceptual,wang2021dual,tan2021legitimate}. A separate category strives to enhance robustness~\cite{kong2020physgan,huang2020universal,hu2022adversarial}. Consequently, a single metric struggles to impartially assess multiple methods.
\end{enumerate}
These problems raise our question: \emph{how does the actual performance of the physical adversarial attacks?} 
In this section, we take the first step in conducting a comprehensive assessment of physical adversarial attack methods, encompassing all the work in this field.

Since the diverse objectives across various works, we summarize six perspectives from existing literature, namely: \textit{Effectiveness}, \textit{Stealthiness}, \textit{Robustness}, \textit{Practicability}, \textit{Aesthetics}, and \textit{Economics}. 
Hingun \textit{et al.}~\cite{hingun2022reap} introduced a large-scale realistic adversarial patch benchmark to assess the effectiveness of attacks.
Parallel to this work,
Wang \textit{et al.}~\cite{wang2023does} evaluated the system-level effect of attack methods in autonomous driving.
{\color{black}Li \textit{et al.}~\cite{li2023towards} proposed an evaluation of visual naturalness.
These evaluation were conducted in settings with a \textit{single perspective} and a \textit{single task}.}
Towards a unified evaluation, we introduce the hexagonal indicator of Physical Adversarial Attack (\textit{hiPAA}) to systematically quantify and compare attack methods across the aforementioned six perspectives.
We define the \textit{hiPAA} as the weighted sum of six components:
\begin{equation}
 \begin{aligned}
hiPAA=& \lambda_1\cdot Eff. + \lambda_2\cdot Rub. + \lambda_3\cdot Ste. \\ 
 &+ \lambda_4\cdot Aes. + \lambda_5\cdot Pra. + \lambda_6\cdot Eco.,
 \end{aligned}
\end{equation}
where the weights of $\lambda_1$ to $\lambda_6$ are assigned based on the importance of each component.
{\color{black}Specifically, we categorized the six components into three tiers based on their importance, assigning different weights to each tier. \textit{Effectiveness} is allocated the highest tier weight of 0.3, followed by \textit{Robustness} and \textit{Stealthiness} with a weight of 0.2 each. The remaining three components are assigned the lowest tier weight of 0.1 each. Overall, the weights for the six dimensions are \{0.3, 0.2, 0.2, 0.1, 0.1, 0.1\}. Their sum equals 1.}

\subsection{Six Perspectives of \textit{hiPAA}}
\noindent{\textbf{Effectiveness.}}
Attack effectiveness is employed to evaluate the influence a method can exert on the victim model.
To evaluate the effectiveness of physical adversarial attacks, our primary concern lies in quantifying the degree of performance degradation induced by these attack methods. Note that while the attack success rate (ASR) serves as a prevalent metric, its applicability across all tasks is not universal. For instance, defining attack success in segmentation tasks can pose inherent challenges. Furthermore, to address variations across different tasks, we compute the percentage of performance degradation:
\begin{equation}
 \begin{aligned}
Eff. = 1-Acc'/Acc,
 \end{aligned}
\end{equation}
where $Acc$ and $Acc'$ represent the model's accuracy without and with attacks, respectively.

\begin{figure*}[t]
  \centering
  \includegraphics[width=0.9\linewidth]{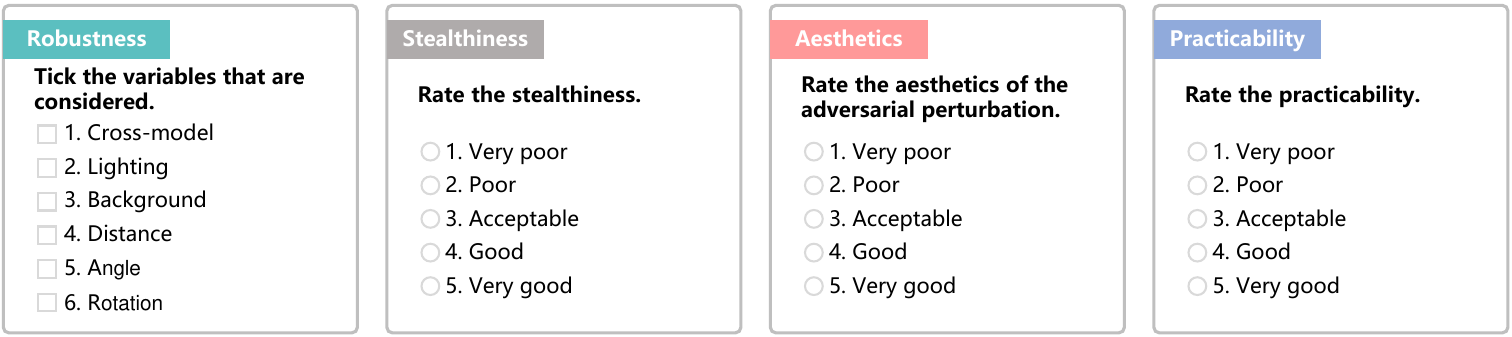}
  \caption{{\color{black}Questions and labels of the four evaluation dimensions design.}}
  \label{figure:hipaaUserStudy}
    \vspace{-0.4cm}
\end{figure*}

\noindent{\textbf{Robustness.}}
Attack robustness is utilized to evaluate the method's stability in dynamic environments.
To evaluate the robustness of physical adversarial attack, we consider three evaluation scenarios: 
1) Cross-model: whether the attack method remains effective when the model changes, 
2) Cross-scenario: whether the attack method can consistently perform in various real-world scenarios, 
and 3) Transformation-resistant: whether the attack method can withstand various real-world transformations, including rotation, camera-to-object distances, view angles, \textit{etc}.
{\color{black}As shown in the ``Robustness'' section of the Fig.~\ref{figure:hipaaUserStudy}, we present six candidate variables for assessing robustness.
TABLE~\ref{tab:personrobustness} serves as an example demonstration of robustness evaluation.}

\noindent{\textbf{Stealthiness.}}
{\color{black}Stealthiness is used to measure the invisibility of adversarial textures or attack methods. In real-world attacks, stealthiness is crucial, as conspicuous attacks are easily detectable and can be thwarted by humans. Since stealthiness is targeted at human observers, we conduct a human study in which participants rate the images using a 5-point Absolute Category Rating (ACR) scale~\cite{li2023towards}.}

\noindent{\textbf{Aesthetics.}}
Aesthetics evaluates the social acceptability of the adversarial textures or attack methods.
This metric is crucial, especially for wearable adversarial mediums~\cite{xu2020adversarial,hu2021naturalistic,hu2022adversarial}. Unusual patterns or appearances may lead users to reject them, whereas solutions that prioritize aesthetics are more likely to be accepted.
Similar to stealthiness, we assess aesthetics through human ratings.

{\color{black}Li \textit{et al.}~\cite{li2023towards} introduced the Dual Prior Alignment (DPA) network to assess the visual naturalness of physical adversarial attacks. 
In this review, we adopt two perspectives, namely stealthiness and aesthetics, to provide a more nuanced evaluation of visual naturalness.
Stealthiness and aesthetics are closely tied to naturalness. 
The distinction between stealthiness and aesthetics lies in the former measuring the invisibility of adversarial textures or attack methods, while the latter evaluates the alignment with human aesthetics under the premise of visible adversarial textures.}

\noindent{\textbf{Practicability.}}
Practicability evaluates the practicality and feasibility of using the attack method in real-world scenarios. It considers factors such as the ease of implementation, availability of resources, and compatibility with existing systems. The higher the practicability, the more feasible and convenient it is to employ the adversarial medium in practical applications.
Similar to stealthiness, we assess practicability through human ratings.

\begin{table*}
    \centering
    \footnotesize
    \caption{
        {{\color{black}Description of the scores for each dimension of the hiPAA indicator.}}
    }
    \label{tab:hipaaDescription}
    \begin{tabular}{p{3cm}p{2.4cm}p{2.7cm}p{8.1cm}}
        \toprule
        Perspective of hiPAA  & Range of values  & Normalization & Description\\
        \midrule
        Effectiveness & [0, 1] & [0, 1] & A higher value indicates a greater decline in model accuracy, reflecting superior attack performance. Conversely, a lower value suggests a lesser decrease in model accuracy, indicating poorer attack performance.\\
        \specialrule{0em}{1pt}{1pt}
        \hline
        \specialrule{0em}{1pt}{1pt}
        Robustness &  \{0, 1, 2, 3, 4, 5, 6\} &  \{0, 0.17, 0.33, 0.50, 0.67, 0.83, 1.00\} & A higher value indicates that the method takes into account a greater number of factors, resulting in better robustness. Conversely, a lower value suggests that the method considers fewer factors, leading to poorer robustness.\\
        \specialrule{0em}{1pt}{1pt}
        \hline
        \specialrule{0em}{1pt}{1pt}
        Stealthiness &  \{1, 2, 3, 4, 5\} &  \{0.2, 0.4, 0.6, 0.8, 1.0\} & A higher value indicates better concealment, while a lower value suggests poorer concealment.\\
        \specialrule{0em}{1pt}{1pt}
        \hline
        \specialrule{0em}{1pt}{1pt}
        Aesthetics &  \{1, 2, 3, 4, 5\} &  \{0.2, 0.4, 0.6, 0.8, 1.0\} & A higher value indicates greater aesthetic appeal, while a lower value suggests less aesthetic appeal.\\
        \specialrule{0em}{1pt}{1pt}
        \hline
        \specialrule{0em}{1pt}{1pt}
        Practicability &  \{1, 2, 3, 4, 5\} &  \{0.2, 0.4, 0.6, 0.8, 1.0\} & A higher value indicates better practicality, while a lower value suggests poorer practicality.\\
        \specialrule{0em}{1pt}{1pt}
        \hline
        \specialrule{0em}{1pt}{1pt}
        Economics & [0, 1]  & [0, 1] & A higher value indicates lower cost, while a lower value suggests higher cost.\\
        \bottomrule
    \end{tabular} 
  \vspace{-0.2cm}
\end{table*}

\noindent{\textbf{Economics.}}
Economics evaluates the resource requirements and expenses associated with implementing and deploying the attack methods. 
Existing methods~\cite{zhu2021fooling,wei2023hotcold} often provide descriptions of the costs associated with manufacturing adversarial mediums.
We assess affordability based on these costs. 
For research that does not explicitly detail these expenses, we make estimations based on experience. 

\subsection{{\color{black}Evaluation Details}}
{\color{black}
For all evaluation perspectives,
TABLE~\ref{tab:hipaaDescription} lists the ranges of values, along with explanations of the score meanings for each dimension.
To ensure fairness, we normalized the scores to fall between 0 and 1.
In Fig.~\ref{figure:hipaaUserStudy}, we present the questions and labels utilized in the human evaluation.
We recruited 30 participants, all of whom were over 18 years old, and provided consent to potentially encounter offensive content during the task.
Before scoring, we provided them with prerequisite knowledge about physical adversarial attacks and evaluation details.
Each participant is required to rate the stealthiness, aesthetics, and practicability of each method according to the provided questions and labels.
}

\subsection{{\color{black}Discussion}}
{\color{black}
While the \textit{hiPPA} offers a valuable means of aggregating multiple performance aspects, it has limitations concerning scenario-specific weighting and cross-task comparison.
Firstly, the \textit{hiPPA} was designed with a specific set of weights to balance various aspects.
However, the importance of these aspects varies across different scenarios. For instance, effectiveness may be prioritized in laboratory settings, while stealthiness and economics might be critical in real-world applications. This variability means that the fixed weights may not always reflect the optimal balance for all use cases. Future work should explore adaptive weighting schemes that dynamically adjust based on scenario-specific requirements or user-defined priorities.

Secondly, the \textit{hiPPA} score's effectiveness component is limited by the diversity of CV tasks, which use different datasets, evaluation protocols, and metrics. Our current approach involves extracting accuracy scores from each paper, leading to inconsistencies due to varying metrics and experimental setups. We mitigate this by normalizing the decrease in accuracy before and after the attack. However, this score should still be interpreted cautiously. Future research should focus on developing task-specific evaluation protocols to facilitate more consistent and comparable assessments of physical adversarial attacks on CV tasks.

}

%% file: 5tasks.tex
{\section{Victim Tasks}
	\label{sec:tasks}}
In this section, we methodically review physical adversarial attacks, categorizing them based on their respective task types. As shown in Fig.~\ref{figure:content}, to our knowledge, there are presently 15 sub-tasks associated with physical adversarial attacks. We organize these tasks into 4 main groups: classification, detection, re-identification, and others. {\color{black}The chronological overview in TABLE~\ref{tab:mediumdefinitions} marks the milestone works in this field.}

{\subsection{Attacks on Classification Tasks}}
Considering the fundamental nature and significant downstream impact, substantial research is concentrated on attacking classification tasks, primarily focusing on general classification and traffic sign classification.
Please refer to Fig.~\ref{figure:generalClassification} and Fig.~\ref{figure:SignClassificationDisplay} for visual representations.

\subsubsection{General Classification}
\begin{figure}[t]
  \centering
  \includegraphics[width=0.98\linewidth]{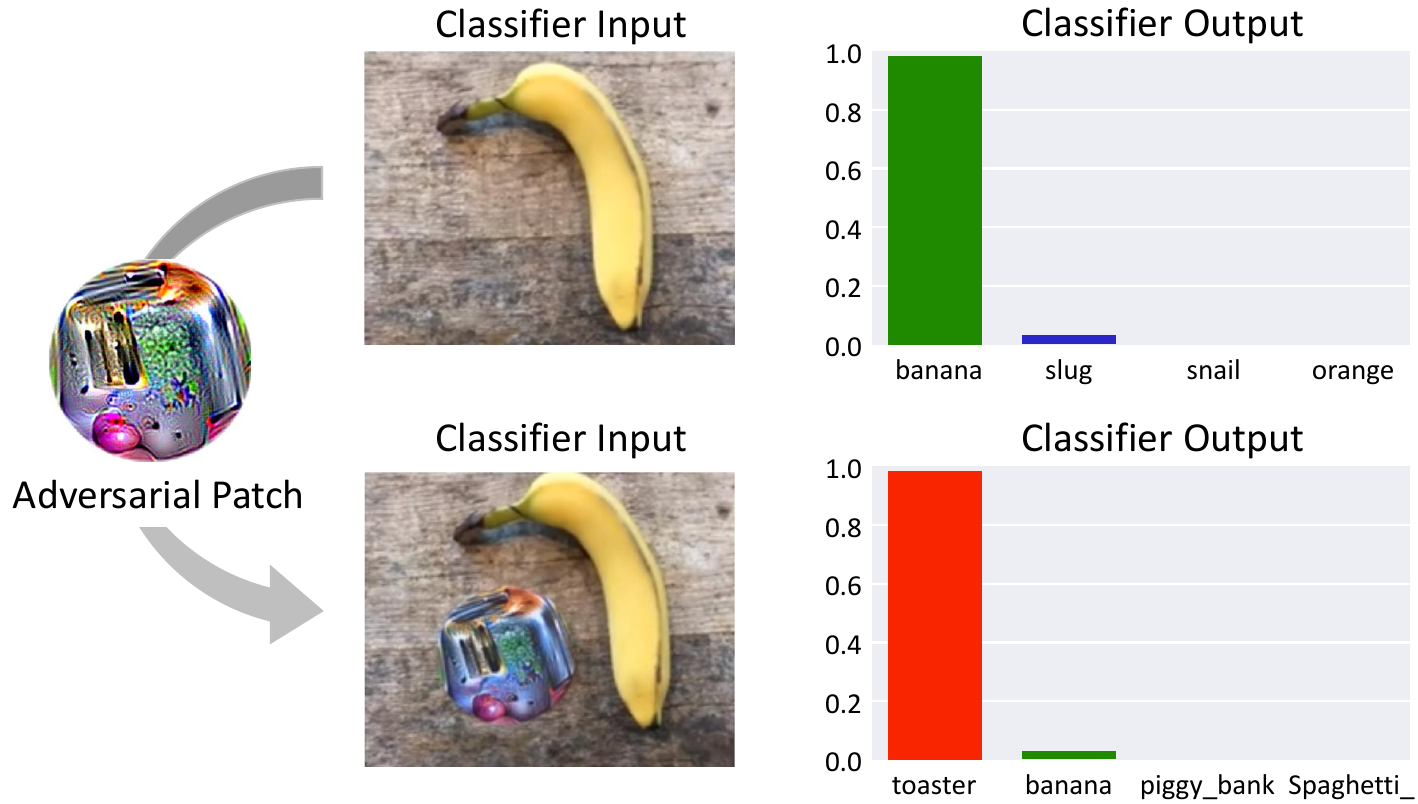}
  \caption{Display of the physical adversarial attack in general classification tasks. Initially, the classifier accurately labels the image as ``banana''. However, when an adversarial patch is placed adjacent to the banana, the classifier misclassifies the image as ``toaster'', despite the continued presence of the banana. Adapted from AdvPatch \cite{brown2017adversarial}.}
  \label{figure:generalClassification}
  \vspace{-0.4cm}
\end{figure}

\begin{figure}
\centering    
\subfigure[A Stop sign with black/white blocks is classified as Speed Limit 45.]{\label{figure:SignClassificationDisplay1}\includegraphics[width=27mm]{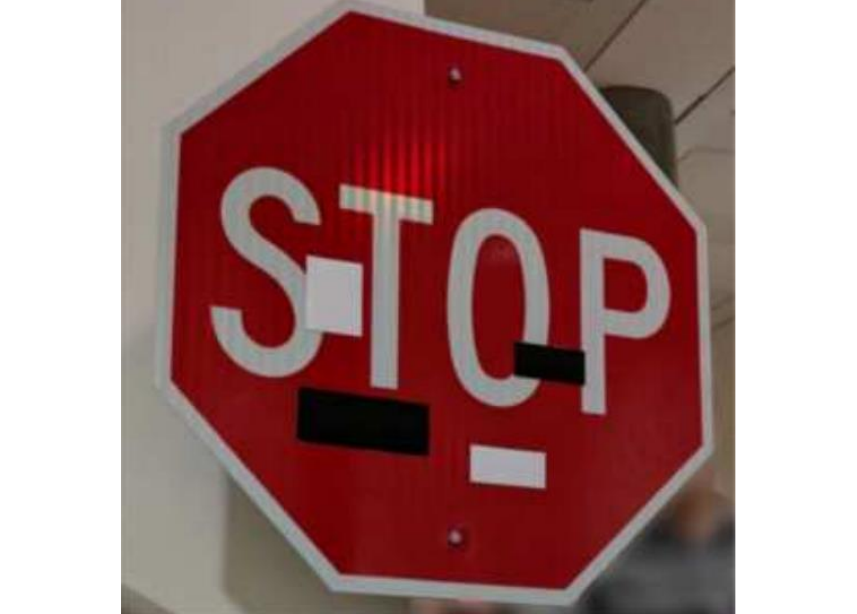}} \hspace{2mm}
\subfigure[A Speed Limit 20 sign with an adversarial patch is classified as Slippery Road.]{\label{figure:SignClassificationDisplay2}\includegraphics[width=27mm]{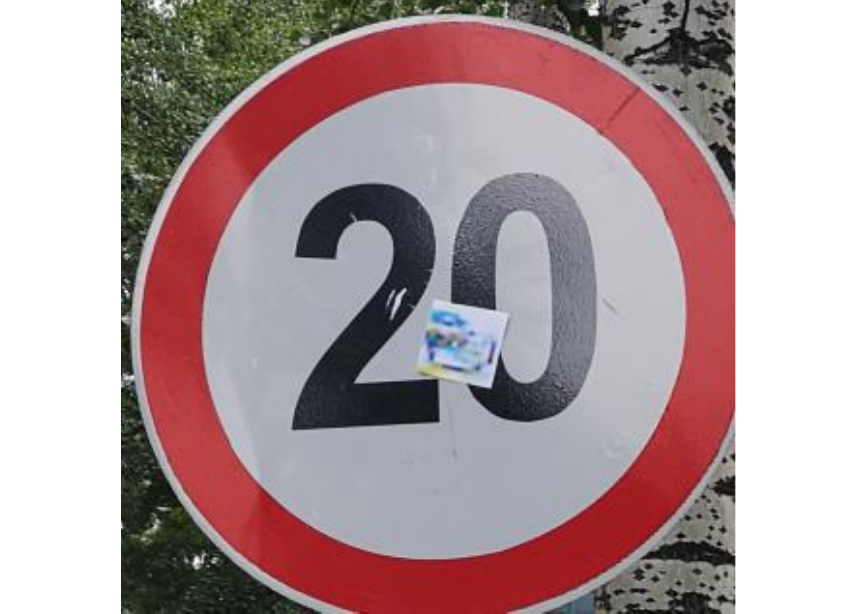}} \hspace{2mm}
\subfigure[A Speed Limit 25 sign with shadows is classified as Speed Limit 35.]{\label{figure:SignClassificationDisplay3}\includegraphics[width=27mm]{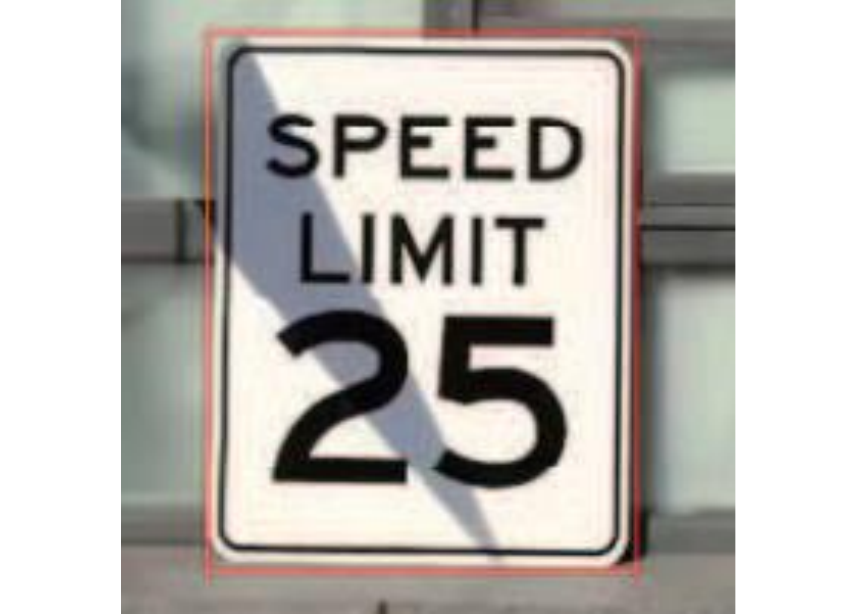}}
\caption{Display of the physical adversarial attack in traffic sign classification tasks. Adapted from RP$_2$~\cite{eykholt2018robust} (a), PS-GAN~\cite{liu2019perceptual} (b), and Adv-Shadow~\cite{zhong2022shadows} (c).}
\label{figure:SignClassificationDisplay}
  \vspace{-0.4cm}
\end{figure}

Brown \textit{et al.}~\cite{brown2017adversarial} proposed the adversarial patch, a milestone in physical adversarial attack methods.
The attacker places a printed circular image patch next to the ``banana'', effectively deceiving the classifier into recognizing the banana as a ``toaster'' with high confidence (see Fig.~\ref{figure:generalClassification}).
Due to its simplicity in production, ease of deployment, and high attack potency, the adversarial patch immediately garnered widespread attention.
Liu \textit{et al.}~\cite{liu2020bias,wang2021universal} developed class-agnostic universal adversarial patches with robust generalization capabilities for attacking classifiers in Automatic Check-out scenarios.
Casper \textit{et al.}~\cite{casper2022robust} designed the ``copy/paste'' attacks, employing adversarial patches to investigate the reliability and interpretability of models.
To unleash the patch-based attack potential, Chen \textit{et al.}~\cite{chen2022shape} designed a Deformable Adversarial Patch (DAPatch) to explore the impact of patch shapes. They simultaneously optimized shape and texture to enhance the attack performance.
While these methods exhibit respectable attack performance, they are prone to visual abnormalities that render them conspicuously noticeable.
To generate a natural-looking patch, Doan \textit{et al.}~\cite{doan2022tnt} proposed searching for naturalistic patches with adversarial effects within the latent space $z$ of Generative Adversarial Networks (GANs)~\cite{goodfellow2020generative}. 

In contrast to patch-based methods that only add perturbations in a limited region, another category of methods adds perturbations across the entire image.
Phy-FGSM~\cite{kurakin2017adversarial} adds carefully designed perturbations to images from the ImageNet dataset~\cite{deng2009imagenet}, prints them out, and then captures them using a cell phone camera. The results demonstrate that the captured images can still successfully attack Inception v3~\cite{szegedy2016rethinking}.
With the generation of digital images and their application in the physical realm, a gap arises between these domains. {\color{black}D2P~\cite{jan2019connecting} was the first to address this domain gap, utilizing conditional GANs~\cite{zhu2017unpaired} to model the digital-to-physical transformation. Its objective is to mitigate the impact of this gap on attack performance.}
MetaAttack~\cite{feng2021meta} and Meta-GAN \cite{feng2023robust} generate robust adversarial examples to maintain attacks in physical dynamics. 
These perturbations are subtle and constrained within a certain range $\varepsilon$, i.e., $\left\| \delta \right\|<\varepsilon$, but they are visible to the human eye, unlike digital adversarial attacks~\cite{su2019one} that are imperceptible.
To make these perturbations appear natural, ABBA~\cite{guo2020watch} disguises the perturbations as motion blur, and AdvCam~\cite{duan2020adversarial} conceals them within natural styles.
In addition to adding perturbations for attacks, ViewFool~\cite{dong2022viewfool} claims that changes in the viewpoint of the target object can also affect the classifier's predictions.
It utilizes adversarial viewpoints to launch attacks and assess the classifier's robustness.

Inspired by One Pixel Attack~\cite{su2019one}, {\color{black}Nichols \textit{et al.}~\cite{nichols2018projecting} pioneered light-based adversarial attacks on classifiers. This method enables attackers to initiate an attack by directing a specially designed light onto the target object.} However, this method has only been validated on low-resolution image datasets CIFAR-10.
Man \textit{et al.}~\cite{man2019poster} manipulated the imaging of printed figures using the ``flare effect'' and the ``blooming effect'' to achieve attacks.
OPAD~\cite{gnanasambandam2021optical} modeled the light emitted by a physical projector to cast adversarial patterns onto target objects, resulting in robust attacks.
Duan \textit{et al.}~\cite{duan2021adversarial} proposed AdvLB, a method that enables the manipulation of the physical parameters of laser beams to execute adversarial attacks, which can be accomplished using handheld laser pointers in real-world scenarios.
To enhance the stealthiness of light-based attacks, Huang \textit{et al.}~\cite{huang2022spaa} devised an optimization algorithm to balance adversarial loss and stealthiness loss. 

As a key component of physical adversarial attack workflow, cameras are responsible for the transformation from the physical scene to a digital image.
In contrast to altering objects in the physical scene, {\color{black}Li \textit{et al.}~\cite{li2019adversarial} innovatively proposed applying translucent stickers directly onto the camera lens to achieve the attack goal.} These specially crafted stickers introduce perturbations into the captured images.
Subsequently, Sayles \textit{et al.}~\cite{sayles2021invisible} also focused on the camera, leveraging the Rolling Shutter Effect inherent in cameras, combined with adversarially illuminating a scene, to introduce an attack signal into the captured images.
Phan \textit{et al.}~\cite{phan2021adversarial} designed adversarial attacks that are effective only under specific camera ISP parameter settings.
These methods provide an alternative approach to implementing physical adversarial attacks by modifying certain camera attributes or settings.

\setlength{\tabcolsep}{4.7pt}
\begin{table}[t]\scriptsize
\centering
\caption{Comparison of the \textit{hiPPA} metric among attacks for the general classification task.
We highlight the minimum and maximum values in \textcolor{blue}{blue} and \textcolor{red}{red}, respectively.}
\vspace{-0.1mm}
\begin{tabular}{lp{0.4cm}p{0.4cm}p{0.4cm}p{0.4cm}p{0.4cm}p{0.4cm}c}
\toprule
 \multirow{2}{*}{Methods} & \multicolumn{6}{c}{Hexagonal Score} & \multirow{2}{*}{hiPAA}\\ \cmidrule(lr){2-7}
                    & Eff. & Rob.  &  Ste.   & Aes.  & Pra.  & Eco. & \\
  \midrule
AdvPatch~\cite{brown2017adversarial} \tiny{NIPS17} & 1.00 & 0.67 & 0.24 & 0.21 & 0.61 & 0.99 & 0.66\\
Phy-FGSM~\cite{kurakin2017adversarial} \tiny{ICLR17} & 0.50 & 0.67 & 0.43 & 0.40 & 0.68  & 0.99&0.58 \\
PTAttack~\cite{nichols2018projecting} \tiny{AAAI18} & 0.78 & 0.50 & 0.64 & 0.66 & 0.83 & 0.95& 0.71\\
EOT~\cite{athalye2018synthesizing} \tiny{PMLR18} & 0.99  & 0.83 & 0.62 & 0.63 & 0.60 & 0.92& {0.80}\\
3DAttack~\cite{zeng2019adversarial}\tiny{CVPR19} & 0.94 & 0.50 & 0.87 & 0.86 & 0.63 & 0.92 & 0.80\\
Poster~\cite{man2019poster} \tiny{S\&P19} & 0.83 & 0.33 & 0.26 & 0.46 & 0.65 & 0.92 & 0.57\\
ACS~\cite{li2019adversarial} \tiny{PMLR19} & 0.49 & 0.33 & 0.88 & 0.87 & 0.85 & 0.99 & 0.66\\
D2P~\cite{jan2019connecting} \tiny{AAAI19} & 0.93 & 0.83 & 0.41 & 0.43 & 0.67 & 0.99 & 0.74\\
AdvACO~\cite{liu2020bias} \tiny{ECCV20} & 0.44 & 0.83 & 0.81 & 0.86 & 0.69 & 0.99& 0.71\\
ABBA~\cite{guo2020watch} \tiny{NIPS20} & 0.85 & 0.50 & 0.69 & 0.64 & 0.61 & 0.99& 0.72\\
AdvCam~\cite{duan2020adversarial} \tiny{CVPR20} & 0.40 & 0.50 & 0.88 & 0.81 & 0.60 & 0.99& 0.64\\
MetaAttack~\cite{feng2021meta} \tiny{ICCV21} & 0.95 & 0.50 & 0.44 & 0.43 & 0.66 & 0.99& 0.68\\
AdvACO+~\cite{wang2021universal} \tiny{TIP21} & 0.72 & 0.83 & 0.89 & 0.81 & 0.69 & 0.99& \textcolor{red}{0.81}\\
InvisPerturb~\cite{sayles2021invisible} \tiny{CVPR21} & 0.94 & 0.67 & 0.81 & 0.43 & 0.45 & 0.00 & 0.67\\
AdvISP~\cite{phan2021adversarial} \tiny{CVPR21} & 0.90 & 0.17 & 0.41 & 0.42 & 0.63 & 0.00& \textcolor{blue}{0.49}\\
OPAD~\cite{gnanasambandam2021optical} \tiny{ICCV21} & 0.43 & 0.50 & 0.28 & 0.48 & 0.88 & 0.85 & {0.51}\\
AdvLB~\cite{duan2021adversarial} \tiny{CVPR21} & 0.88 & 0.67 & 0.85 & 0.41 & 0.81 & 0.92 & 0.78\\
CPAttack~\cite{casper2022robust} \tiny{NIPS22} & 1.00 & 0.17 & 0.26 & 0.26 & 0.63 & 0.99 & 0.57\\
TnTAttack~\cite{doan2022tnt} \tiny{TIFS22} & 0.95 & 0.67 & 0.43 & 0.83 & 0.67 & 0.99 & 0.75\\
DAPatch~\cite{chen2022shape} \tiny{ECCV22} & 0.44 & 0.67 & 0.26 & 0.27 & 0.64 & 0.99 & 0.51\\
ViewFool~\cite{dong2022viewfool} \tiny{NIPS22} & 0.92 & 0.50 & 1.00 & 0.89 & 0.43 & 0.99 & \textcolor{red}{0.81}\\
SPAA~\cite{huang2022spaa} \tiny{VR22} & 1.00 & 0.33 & 0.23 & 0.41 & 0.89 & 0.92& 0.63\\
Meta-GAN \cite{feng2023robust} \tiny{TIFS23} & 0.95 & 0.67 & 0.44 & 0.43 & 0.61 & 0.99 & 0.71\\
{\color{black}TT3D~\cite{huang2024towards} \tiny{CVPR24}} & 0.83 & 0.83 & 0.75 & 0.57 & 0.70 & 0.86 & 0.78 \\
\bottomrule
\end{tabular}
\label{tab:classifier}
  \vspace{-0.4cm}
\end{table}

Apart from utilizing 2D images, methods involving 3D objects have also been explored. {\color{black}Athalye \textit{et al.}~\cite{athalye2018synthesizing} fabricated the first 3D adversarial object—a turtle misclassified as a rifle from multiple viewpoints. Their proposed Expectation Over Transformation (EOT) technique enhances the robustness of adversarial examples, later widely applied in physical adversarial attack domains.}
Zeng \textit{et al.}\cite{zeng2019adversarial} perturbed 3D physical properties under a renderer to deceive 3D object classification and visual question-answering models. {\color{black}Subsequently, aiming to enhance cross-task transferability of 3D adversarial examples, Huang \textit{et al.}\cite{huang2024towards} introduced TT3D, enabling attacks across classification, detection, and captioning tasks.}

The evaluation results of attacks on general classification tasks are summarized in TABLE~\ref{tab:classifier}. These results indicate that the performance of these attack methods has not continuously improved over the years. The lack of a comprehensive evaluation framework has contributed to this issue. For example, while CPAttack~\cite{casper2022robust} performs well in terms of effectiveness and economics, it neglects robustness.

\subsubsection{Traffic Sign Classification}
\begin{figure}
\centering    
\subfigure[RP$_2$~\cite{eykholt2018robust}]{\label{figure:SignClassificationEval1}\includegraphics[width=29mm]{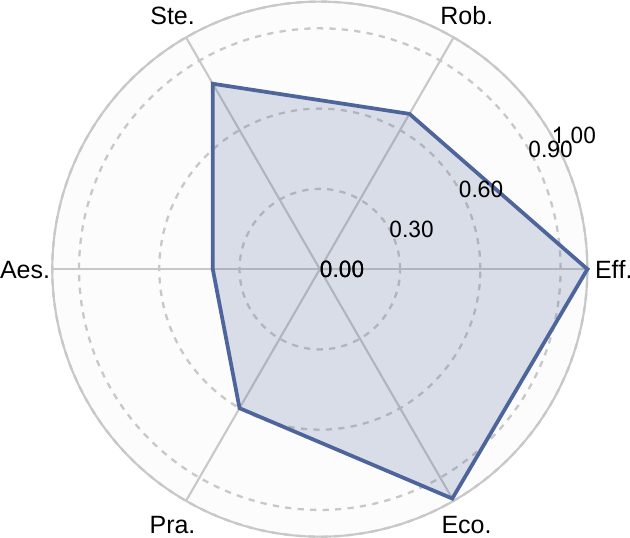}} \hspace{-1mm}
\subfigure[PS-GAN~\cite{liu2019perceptual}]{\label{figure:SignClassificationEval2}\includegraphics[width=29mm]{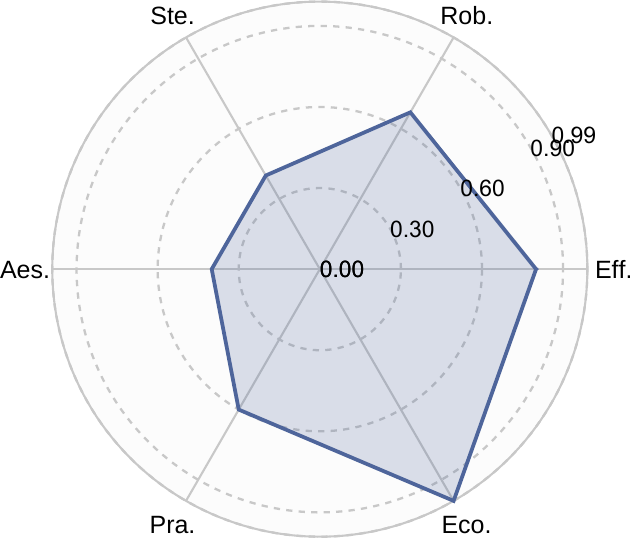}} \hspace{-1mm}
\subfigure[Adv-Shadow~\cite{zhong2022shadows}]{\label{figure:SignClassificationEval3}\includegraphics[width=29mm]{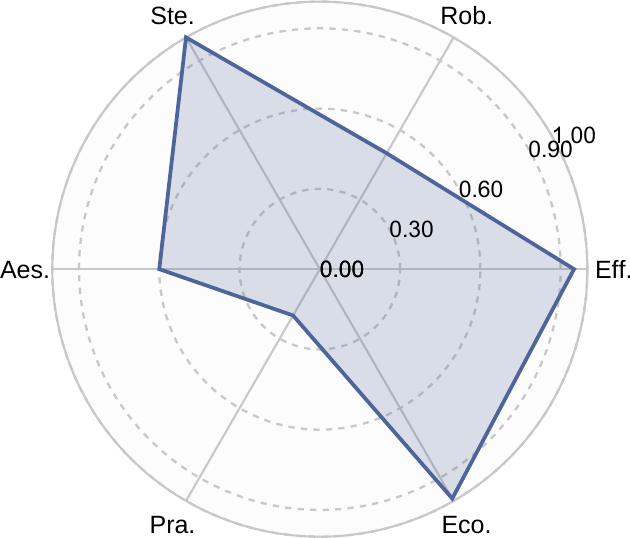}}
\caption{Comparison of six perspectives of hiPAA across three physical adversarial attack methods on traffic sign classification task.}
\label{figure:SignClassificationEval}
  \vspace{-0.4cm}
\end{figure}

The classification of traffic signs plays a pivotal role in aiding autonomous driving systems to comprehend scenes and make informed decisions~\cite{wang2022defensive}. Consequently, safety assessments in this domain have garnered considerable attention (see Fig.~\ref{figure:SignClassificationDisplay}).
{\color{black}Eykholt \textit{et al.}~\cite{eykholt2018robust} were the first to investigate the robustness of traffic sign classifiers and introduced Robust Physical Perturbations (RP$_2$), a method that misguides these classifiers by affixing black and white blocks onto signs.}
Liu \textit{et al.}~\cite{liu2019perceptual} proposed a perceptual-sensitive generative adversarial network (PS-GAN), which generates adversarial patches that visually resemble the scrawls typically found on signs in the real world, enhancing their stealthiness.
Subsequently, building on potential real-world interferences, Zhong \textit{et al.}~\cite{zhong2022shadows} subsequently introduced Adv-Shadow to evaluate the influence of shadows on traffic sign classifiers. {\color{black}Similarly, Wang \textit{et al.}~\cite{wang2023rfla} proposed using reflections from mirrors to launch attacks. Experimental results show that optimized shadows and reflections can effectively deceive target models.}
Fig.~\ref{figure:SignClassificationEval} presents a comparison of RP$_2$, PS-GAN, and Adv-Shadow in terms of the six perspectives of the \textit{hiPPA} metric.
{\color{black}Results show that such attacks demonstrate excellence in effectiveness and economics, but they lack aesthetics and practicality. These observations provide potential directions for further improvement.}

\begin{figure}
\centering    
\subfigure[Vehicle Detection]{\label{figure:cardet}\includegraphics[width=29mm]{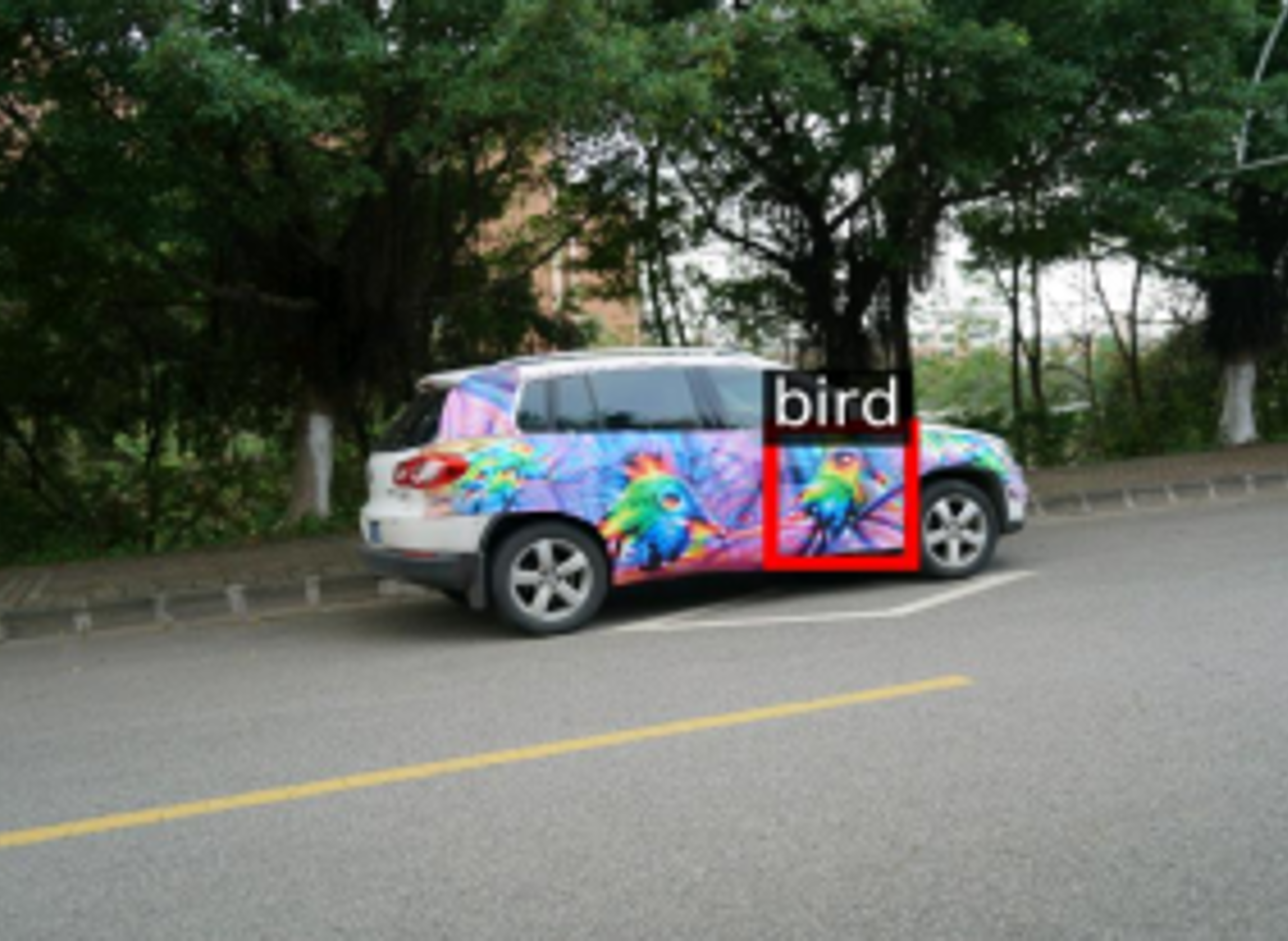}} \hspace{-1mm}
\subfigure[Person Detection]{\label{figure:persondet}\includegraphics[width=29mm]{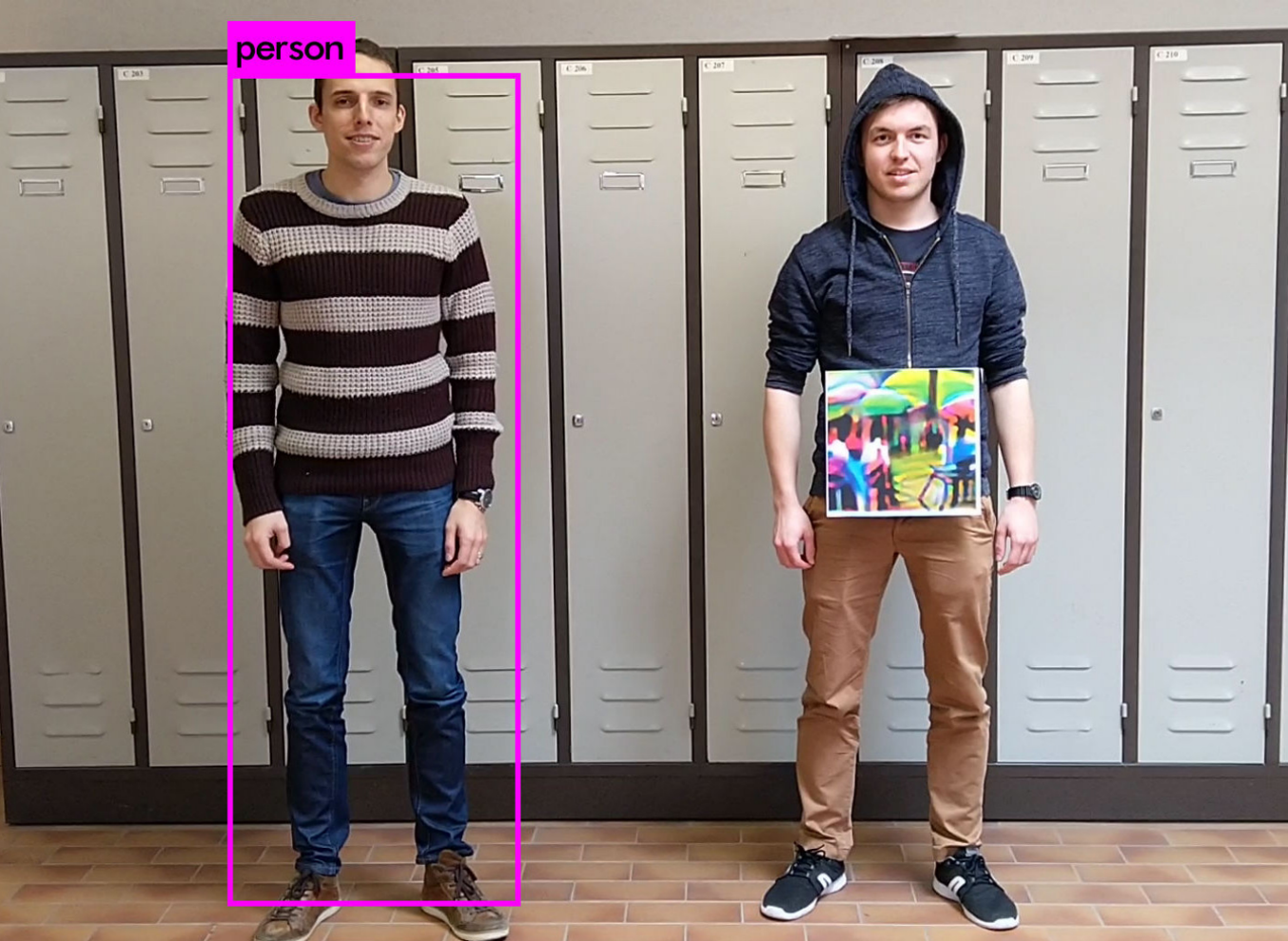}} \hspace{-1mm}
\subfigure[Sign Detection]{\label{figure:signdet}\includegraphics[width=29mm]{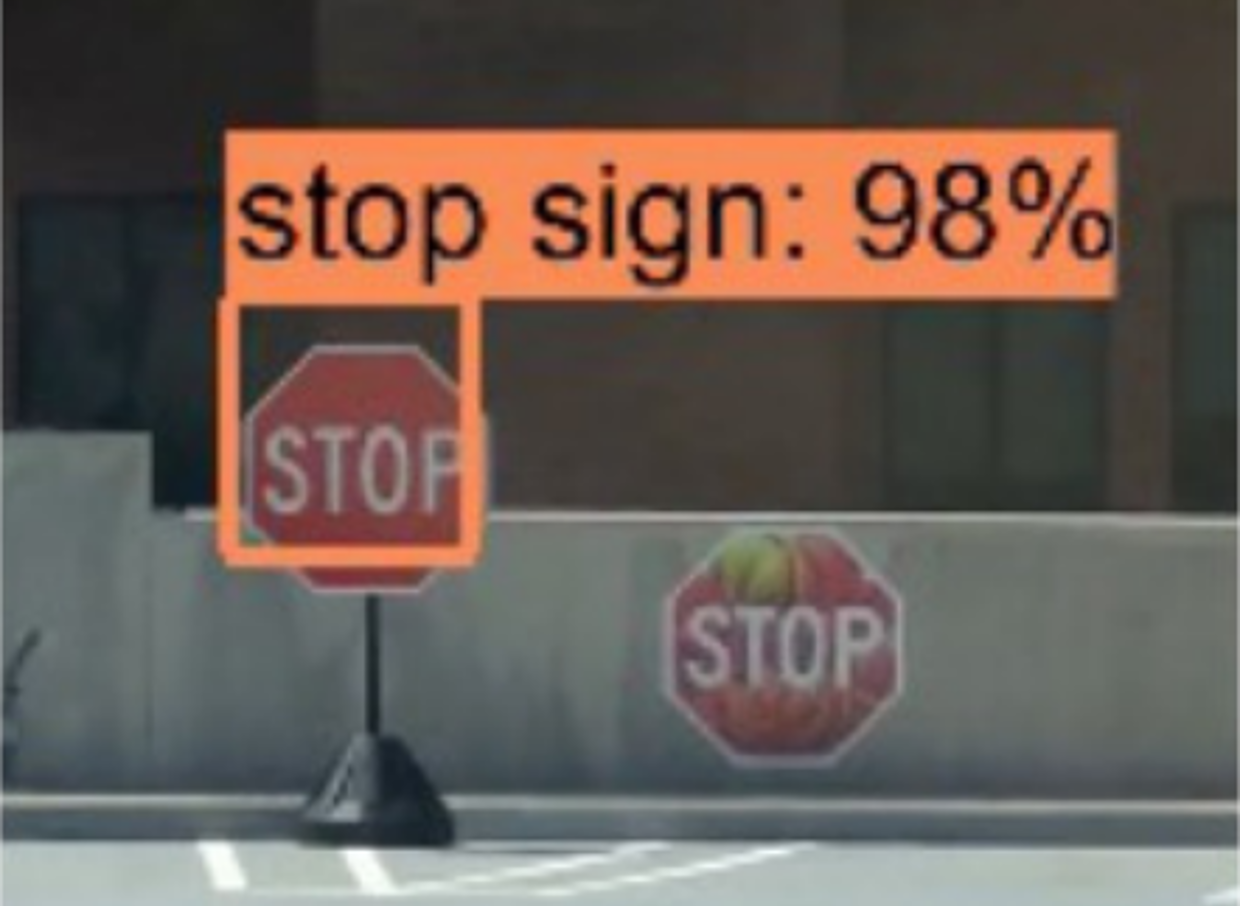}}
\caption{Display of the physical adversarial attack in vehicle, person, and traffic sign detection tasks. The detectors fail to detect the perturbed target. Adapted from UPC~\cite{huang2020universal} (a), AdvYOLO~\cite{thys2019fooling} (b), and ShapeShifter~\cite{chen2018shapeshifter} (c).}
\label{figure:detection}
  \vspace{-0.4cm}
\end{figure}

{\subsection{Attacks on Detection Tasks}
	\label{sec:detection}}
Physical adversarial attacks on detectors primarily focus on evading detection.
As shown in Fig.~\ref{figure:detection}, we display examples of attacking the vehicle, person, and sign detection tasks. 

\subsubsection{Vehicle Detection}
Physical adversarial attacks on vehicle detection focus on applying distinctive patterns to a vehicle's exterior to evade detection models.

{\color{black}Zhang \textit{et al.}~\cite{zhang2018camou} were the first to effectively camouflage a car in a 3D simulation environment.} They employed the photorealistic Unreal Engine 4\footnote{\url{https://www.unrealengine.com/}} to simulate the complex transformations induced by the physical environment comprehensively. This engine offers a comprehensive set of configuration parameters, including camouflage resolution, patterns, 3D vehicle models, camera settings, and environmental variables.
Similar to this work, Wu \textit{et al.}~\cite{wu2020physical} employed the open-source simulator CARLA~\cite{dosovitskiy2017carla}. 
They introduced an Enlarge-and-Repeat process and a discrete search method to craft physically adversarial textures. 
In addition, Duan \textit{et al.}~\cite{duan2022learning} utilized the Unity\footnote{\url{https://unity.com/}} and introduced a method called Coated Adversarial Camouflage (CAC). 

The neural renderer is commonly used in 2D-to-3D transformation. One of the applications is to wrap the texture image to the 3D model, which then is rendered to the 2D image~\cite{kato2018neural,thies2019deferred,rematas2020neural}. Thus utilizing the neural renderer to paint the adversarial stickers onto the vehicle surface is being pervasively used. Full-coverage Camouflage Attack (FCA)~\cite{wang2022fca} tries rendering the non-planar texture over the full vehicle surface to overcome the partial occluded and long-distance issues. It bridges the gap between digital attacks and physical attacks via a differentiable neural renderer. Then, FCA introduces a transformation function to transfer the rendered camouflaged vehicle into a photo-realistic scenario. It outperforms other advanced attacks and achieves higher attack performance on both digital and physical attacks.

However, existing neural renderers cannot fully represent various real-world transformations due to a lack of control of scene parameters compared to legacy photo-realistic renderers. 
Motivated by the challenge, 
Suryanto \textit{et al.}~\cite{suryanto2022dta} presented the Differentiable Transformation Attack (DTA), a framework for generating effective physical adversarial camouflage on 3D objects. It combines the advantages of a photorealistic rendering engine with the differentiability of a novel rendering technique, outperforming previous works in both effectiveness and transferability.

\begin{figure}[t]
  \centering
  \includegraphics[width=0.99\linewidth]{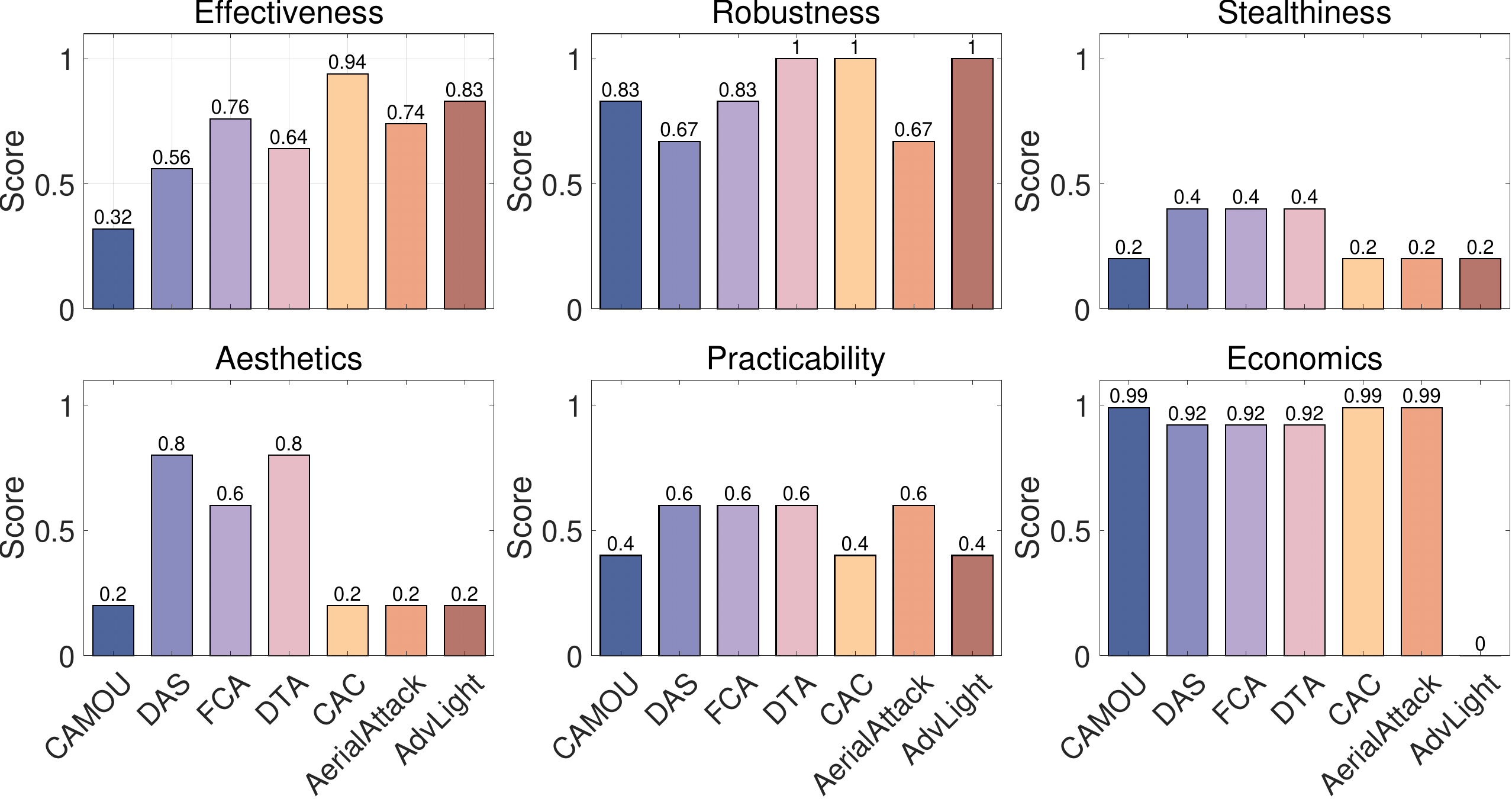}
  \caption{{\color{black}Comparison of physical adversarial attack methods on vehicle detection tasks. These methods encompass CAMOU~\cite{zhang2018camou}, DAS~\cite{wang2021dual}, FCA~\cite{wang2022fca}, DTA~\cite{suryanto2022dta}, CAC~\cite{duan2022learning}, AerialAttack~\cite{du2022physical}, and AdvLight~\cite{10095895}.}}
  \label{figure:lidarone}
  \vspace{-0.4cm}
\end{figure}

The evaluation results of attack methods on vehicle detection tasks are summarized in Fig.~\ref{figure:lidarone}. From these results, the following insights can be drawn.
First, all attack methods do not perform ideally in terms of stealthiness. This is due to the fact that vehicles are large entities with a single color, and any external alterations are likely to be noticeable. 
Second, no method excels in all six aspects; each has its own strengths. For instance, AdvLight~\cite{10095895} demonstrates strong robustness but fares poorly in terms of economics.

\subsubsection{Person Detection}
The objective of physical adversarial attacks on person detection is to conceal a person from detection models in the real world.
Refer to TABLE~\ref{tab:personsign} for the \textit{hiPPA} evaluation, and TABLE~\ref{tab:personrobustness} presents the robustness evaluation.

Yang \textit{et al.}~\cite{yang2018building} were the first to propose attacking person detectors in the real world.
The adversarial patches they generated caused the accuracy of the Tiny YOLO detector~\cite{redmon2016you} to drop from 1.00 to 0.28. 
Thys \textit{et al.}~\cite{thys2019fooling} designed a small (40cm $\times$ 40cm) adversarial patch that, when held by an attacker, can deceive the YOLOv2~\cite{redmon2017yolo9000}. 
Given an input image, mainstream detectors predict the bounding boxes ${{\mathcal{V}}{pos}}$, the object probability ${{\mathcal{V}}{obj}}$, and the class score ${{\mathcal{V}}{cls}}$. During the training phase, they minimize ${{\mathcal{V}}{obj}}$ and ${{\mathcal{V}}_{cls}}$ to cause the detector to ignore persons. Meanwhile, the INRIAPerson dataset~\cite{dalal2005histograms} provides person instances that facilitate the generation of effective adversarial patches.
Subsequently, {\color{black}T-SEA~\cite{huang2022t} achieves high attack transferability through a series of strategies that involve self-ensembling the input data, the victim model, and the adversarial patch.}

Not being satisfied with attacking detectors with printed cardboard, Xu \textit{et al.}~\cite{xu2020adversarial} crafted T-shirts with the generated adversarial patches. We termed their approach AdvT-shirt. {\color{black}Due to the non-rigid deformation of a T-shirt caused by pose changes of a moving person, AdvT-shirt employs a TPS-based transformer to ensure the attack's efficacy in the physical world.} Parallel to this work, Wu \textit{et al.}~\cite{wu2020making} made a wearable invisibility cloak that, when placed over an object either digitally or physically, makes that object invisible to detectors. 
To fairly evaluate the effectiveness of different physical attacks, Huang \textit{et al.}~\cite{huang2020universal} presented the first standardized dataset, AttackScenes, which simulates the real 3D world under controllable and reproducible settings to ensure that all experiments are conducted under fair comparisons for future research in this domain. In addition, they proposed the Universal Physical Camouflage (UPC) attack, which crafts adversarial patterns by jointly fooling the region proposal network, as well as misleading the classifier and the regressor to output errors. 

Legitimate Adversarial Patches (LAP)~\cite{tan2021legitimate} and DAC~\cite{sun2023differential} aim to evade both human perception and detection models in real-world settings. LAP generates cartoon-style adversarial patches, while DAC integrates background styles into adversarial textures to balance effectiveness and stealthiness.
Generative adversarial networks (GANs) have the ability to efficiently generate desired samples~\cite{wang2021generative}. Considering this, NAP \cite{hu2021naturalistic} crafts adversarial patches by leveraging the learned image manifold of BigGAN~\cite{brock2018large} and StyleGAN~\cite{karras2021alias} pretrained on real-world images. Moreover, in this work, the MPII Human Pose dataset~\cite{andriluka20142d} is introduced to provide the diversity of training data. 

{\color{black}To address the segment-missing problem and support multi-angle attacks, Hu \textit{et al.}~\cite{hu2022adversarial} introduced a novel generative method, the Toroidal-Cropping-based Expandable Generative Attack (TC-EGA).} This approach generates adversarial textures characterized by repetitive structures.
Unlike the prior studies, in the first stage, TC-EGA trains a fully convolutional network (FCN)~\cite{long2015fully,springenberg2014striving} as the generator to produce textures by sampling random latent variables as input. In the second stage, TC-EGA searches for the best local pattern of the latent variable with a cropping technique: Toroidal Cropping~\cite{hatcher2005algebraic}. 
They have produced a variety of clothing items, such as T-shirts, skirts, and dresses, that have been printed with generated adversarial textures. These clothing items are effective for carrying out attacks when the wearer turns around or changes their posture.
In the subsequent year, Hu \textit{et al.}~\cite{Hu_2023_CVPR} 
improved the attack's stealthiness by refining the clothing texture based on the TC-EGA method. Specifically, they introduced camouflage textures with a high naturalness score, resulting in the production of natural-looking adversarial clothing.

The security of DNNs-based models has received substantial attention in the context of visible light, but its exploration in thermal infrared imaging remains incomplete. Additionally, thermal infrared detection systems hold significant relevance in various security-related domains, such as autonomous driving~\cite{martin2019drive} and night surveillance~\cite{ye2021dynamic}.
{\color{black}Thus, Zhu \textit{et al.}~\cite{zhu2021fooling} presented a groundbreaking method, named AdvBulbs, which is the first to enable physical attacks on thermal infrared person detectors.} 
Their design is classified as a patch-based attack, with experiments conducted on the Teledyne FLIR ADAS Thermal dataset~\cite{FLIR}. The following year, Zhu \textit{et al.}~\cite{zhu2022infrared} designed infrared invisible clothing based on a new material aerogel.
Compared with the AdvBulbs, infrared invisible clothing hid from infrared detectors from multiple angles. 
Since then, this field has garnered significant attention.
Wei \textit{et al.}~\cite{Wei_2023_CVPR} also used aerogel material, with the difference that they designed irregularly shaped patches for the attack. 
Parallel to this work, Wei \textit{et al.}~\cite{wei2023hotcold} ingeniously employed anti-fever stickers and heating pads to create adversarial patches. Instead of optimizing the texture and structural characteristics of the patches, their emphasis lies in studying the impact of patch size, shape, and placement on the attacks.
Wei \textit{et al.}~\cite{wei2023unified} introduced the concept of cross-modal physical adversarial attacks, which remain effective under both thermal infrared and visible light imaging modalities.

\setlength{\tabcolsep}{4.5pt}
\begin{table}[t]\scriptsize
\centering
\caption{Comparison of the \textit{hiPPA} metric among attack methods for both the person detection task and traffic sign detection task. We highlight the minimum and maximum values using \textcolor{blue}{blue} and \textcolor{red}{red}, respectively.}
\vspace{-0.1mm}
 \begin{threeparttable}
\begin{tabular}{lp{0.4cm}p{0.4cm}p{0.4cm}p{0.4cm}p{0.4cm}p{0.4cm}c}
\toprule
 \multirow{2}{*}{Methods} & \multicolumn{6}{c}{Hexagonal Score} & \multirow{2}{*}{hiPAA}\\ \cmidrule(lr){2-7}
                    & Eff. & Rob.  &  Ste.   & Aes.  & Pra.  & Eco. & \\
  \midrule
  InvisibleCloak~\cite{yang2018building} \tiny{UEMCON18} & 0.72 & 0.67 & 0.25 & 0.25 & 0.61 & 0.29 & \textcolor{blue}{0.52} \\
  AdvYOLO~\cite{thys2019fooling} \tiny{CVPRW19} & 0.75 & 0.50 & 0.25 & 0.24 & 0.62 & 0.99& 0.56\\
  AdvT-shirt~\cite{xu2020adversarial} \tiny{ECCV20}  & 0.43 & 0.67 & 0.67 & 0.68 & 0.81 & 0.95& 0.64\\
  UPC~\cite{huang2020universal} \tiny{CVPR20} & 0.93 & 0.67 & 0.26 & 0.24 & 0.81 & 0.91& 0.66\\
  AdvCloak~\cite{wu2020making} \tiny{ECCV20} & 0.50 & 0.83 & 0.61 & 0.60 & 0.81 & 0.95& 0.67\\
  NAP~\cite{hu2021naturalistic} \tiny{ICCV21} & 0.66 & 0.50 & 0.85 & 0.88 & 0.81 & 0.95& 0.73\\
  LAP~\cite{tan2021legitimate} \tiny{ACM MM21} & 0.52 & 0.67 & 0.88 & 0.89 & 0.81 & 0.95& 0.73\\
  AdvBulbs~\cite{zhu2021fooling} \tiny{AAAI21} & 0.65 & 0.83 & 0.23 & 0.43 & 0.68 & 0.99& 0.62\\
  TC-EGA~\cite{hu2022adversarial} \tiny{CVPR22} & 0.65 & 1.00 & 0.41 & 0.28 & 0.81 & 0.95& 0.68\\
  InvisClothing~\cite{zhu2022infrared} \tiny{CVPR22} & 0.88 & 1.00 & 0.27 & 0.27 & 0.27 & 0.00& 0.57\\
  AdvInfrared~\cite{Wei_2023_CVPR} \tiny{CVPR23}  & 0.85 & 0.50 & 1.00 & 0.25 & 0.65 & 0.95& 0.74\\
  T-SEA~\cite{huang2022t} \tiny{CVPR23}  & 0.99 & 0.67 & 0.26 & 0.24 & 0.69 & 0.29& 0.60\\
  \color{black}{DAC~\cite{sun2023differential} \tiny{NN23}} & 0.75 & 0.67 & 0.63 & 0.37 & 0.00 & 0.93 & 0.61\\
  AdvCaT~\cite{Hu_2023_CVPR} \tiny{CVPR23} & 0.85 & 1.00 & 0.86 & 0.87 & 0.81 & 0.91& \textcolor{red}{0.89}\\
  HOTCOLD Block~\cite{wei2023hotcold} \tiny{AAAI23}  & 0.57 & 0.67 & 1.00 & 0.27 & 0.65 &0.99 & 0.70\\
  CMPatch~\cite{wei2023unified} \tiny{ICCV23} & 0.62 & 0.67 & 1.00 & 0.26 & 0.81 & 0.95 & 0.72\\
  {\color{black}IAPatch~\cite{wei2023infrared} \tiny{IJCV23}} & 0.88 & 0.50 & 1.00 & 0.26 & 0.81 &  0.95 & 0.77 \\
 \midrule
    ShapeShifter~\cite{chen2018shapeshifter} \tiny{EP18} & 0.93 & 1.00 & 0.26 & 0.24 & 0.63 &0.99& 0.72\\  
    RP$_2$+~\cite{220580Song} \tiny{USENIX18} & 0.85 & 0.67 & 0.81 & 0.41 & 0.67 &0.99& 0.76\\  
    NestedAE~\cite{zhao2019seeing} \tiny{CCS19} & 0.92 & 0.67 & 0.26 & 0.25 & 0.66 &0.99& 0.65\\
    LPAttack~\cite{yang2020beyond}\tiny{AAAI20} & 0.80 & 1.00 & 0.27 &0.40 &0.47 &0.99& 0.68\\  
    TransPatch~\cite{zolfi2021translucent} \tiny{CVPR21} & 0.42 & 0.33 & 0.88 & 0.84 & 0.85 &0.99& \textcolor{blue}{0.64}\\
    AdvMarkings~\cite{272270} \tiny{USENIX21} & 1.00 & 0.83 & 1.00 & 0.64 & 0.89 &0.99& \textcolor{red}{0.92}\\
    SLAP~\cite{272218}\tiny{USENIX21} & 0.99 & 1.00 & 0.25 & 0.26  & 0.65 &0.92& 0.73\\  
    AITP~\cite{sava2022assessing} \tiny{AISec22} & 0.90 & 0.83 & 0.27  & 0.46 & 0.61 &0.99& 0.70\\  
    AdvLS~\cite{hu2023adversarial}\tiny{PMLR23} & 0.66 & 0.50 & 0.86 & 0.42 & 0.88 &0.92& 0.69\\  
    {\color{black}TPatch~\cite{zhu2023tpatch} \tiny{USENIX23}} & 1.00 & 0.83 & 0.77 & 0.83 & 0.51 & 0.57 & 0.81\\
\bottomrule
\end{tabular}
 \end{threeparttable}
\label{tab:personsign}
  \vspace{-0.4cm}
\end{table}

\subsubsection{Traffic Sign Detection}
Traffic sign detection involves identifying road signs (e.g., stop signs and lane lines) in driving scenarios, essential for autonomous driving. Due to its critical role in safety and decision-making, adversarial attacks on traffic sign detection have evolved to improve the resilience of detection algorithms. The comparative results of these methods are shown at the bottom of TABLE~\ref{tab:personsign}.

The mainstream detectors prune the region proposals by using heuristics like non-maximum suppression (NMS)~\cite{ren2015faster,yolov52013}. The pruning operations are usually non-differentiable. However, generating adversarial perturbations pervasively requires calculating a backward gradient end to end. The non-differentiable operations make it hard to optimize the objective loss. To tackle this problem, Chen \textit{et al.}~\cite{chen2018shapeshifter} carefully studied the Faster R-CNN object detector~\cite{ren2015faster} and successfully performed optimization-based attacks using gradient descent and backpropagation. Concretely, they ran the forward pass of the region proposal network and fixed the pruned region proposals as fixed constants to the second stage classification in each iteration. 
The RP$_2$ algorithm of \cite{eykholt2018robust} only focuses on attacking the traffic sign classification task. Following this line, Song \textit{et al.}~\cite{220580Song} extended the RP$_2$ to provide proof-of-concept attacks for object detection networks. 
They experimented with the YOLOv2~\cite{redmon2017yolo9000} and achieved 85.6\% ASR in an indoor environment and 72.5\% ASR in an outdoor environment.

\setlength{\tabcolsep}{4.5pt}
\begin{table}[t]\scriptsize
\centering
\caption{Evaluation of the robustness dimension of the \textit{hiPPA} metric on attacking person detection task.}
\vspace{-0.1mm}
 \begin{threeparttable}
\begin{tabular}{cp{0.4cm}p{0.4cm}p{0.4cm}p{0.4cm}p{0.4cm}lp{0.4cm}}
\toprule
  \multirow{3}{*}{\makecell[c]{Cross\\-model}}&   \multicolumn{2}{c}{Cross-scenario} & \multicolumn{3}{c}{Transformation}& \multirow{3}{*}{Method} & \multirow{3}{*}{Rob.} \\
  \cmidrule(lr){2-3} \cmidrule(lr){4-6}
    & \textit{Lig.} & \textit{Bac.}  & \textit{Dis.} & \textit{Ang.}  & \textit{Rot.} &   &   \\
  \midrule
  \usym{2717} & \usym{1F5F8} & \usym{1F5F8} & \usym{1F5F8} & \usym{2717} & \usym{1F5F8} & InvisibleCloak~\cite{yang2018building} & 0.67 \\
  \usym{2717} & \usym{1F5F8} & \usym{1F5F8} & \usym{2717} & \usym{2717} & \usym{1F5F8} & AdvYOLO~\cite{thys2019fooling} & 0.50 \\
  \usym{2717} & \usym{1F5F8} & \usym{1F5F8} & \usym{1F5F8} & \usym{2717} & \usym{1F5F8} & AdvT-shirt~\cite{xu2020adversarial} & 0.67  \\
  \usym{1F5F8} & \usym{1F5F8} & \usym{1F5F8} & \usym{1F5F8} & \usym{2717} & \usym{2717} & UPC~\cite{huang2020universal} & 0.67 \\
  \usym{1F5F8} & \usym{1F5F8} & \usym{1F5F8} & \usym{1F5F8} & \usym{2717} & \usym{1F5F8} & AdvCloak~\cite{wu2020making} & 0.83 \\
  \usym{1F5F8} & \usym{2717} & \usym{1F5F8} & \usym{2717} & \usym{2717} & \usym{1F5F8} & NAP~\cite{hu2021naturalistic} & 0.50 \\
  \usym{2717} & \usym{1F5F8} & \usym{1F5F8} & \usym{1F5F8} & \usym{2717} & \usym{1F5F8} &LAP~\cite{tan2021legitimate} & 0.67  \\
  \usym{1F5F8} & \usym{1F5F8} & \usym{1F5F8} & \usym{1F5F8} & \usym{2717} & \usym{1F5F8} &AdvBulbs~\cite{zhu2021fooling} & 0.83 \\
  \usym{1F5F8} & \usym{1F5F8} & \usym{1F5F8} & \usym{1F5F8} & \usym{1F5F8} & \usym{1F5F8} &TC-EGA~\cite{hu2022adversarial} & 1.00 \\
  \usym{1F5F8} & \usym{1F5F8} & \usym{1F5F8} & \usym{1F5F8} & \usym{1F5F8} & \usym{1F5F8} &InvisClothing~\cite{zhu2022infrared} & 1.00 \\
  \usym{2717} & \usym{1F5F8} & \usym{1F5F8} & \usym{1F5F8} & \usym{2717} & \usym{2717} & AdvInfrared~\cite{Wei_2023_CVPR} & 0.50  \\
  \usym{1F5F8} & \usym{1F5F8} & \usym{1F5F8} & \usym{2717} & \usym{2717} & \usym{1F5F8} & T-SEA~\cite{huang2022t} & 0.67  \\
  \usym{1F5F8} & \usym{2717} & \usym{1F5F8} & \usym{1F5F8} & \usym{2717} & \usym{1F5F8} &\color{black}{DAC~\cite{sun2023differential}} & 0.67 \\
  \usym{1F5F8} & \usym{1F5F8} & \usym{1F5F8} & \usym{1F5F8} & \usym{1F5F8} & \usym{1F5F8} &AdvCaT~\cite{Hu_2023_CVPR} & 1.00 \\
  \usym{1F5F8} & \usym{1F5F8} & \usym{1F5F8} & \usym{1F5F8} & \usym{2717} & \usym{2717} &HOTCOLD Block~\cite{wei2023hotcold} & 0.67  \\
  \usym{1F5F8} & \usym{1F5F8} & \usym{1F5F8} & \usym{1F5F8} & \usym{2717} & \usym{2717} &CMPatch~\cite{wei2023unified} & 0.67 \\
 \bottomrule
\end{tabular}
        \scriptsize{$\bullet$ \textit{Lig.}, \textit{Bac.}, \textit{Dis.}, \textit{Ang.}, and \textit{Rot.} represent lighting, background, camera-to-object distances, view angles, and rotation respectively.}\\
 \end{threeparttable}
\label{tab:personrobustness}
  \vspace{-0.4cm}
\end{table}

\begin{figure*}
\centering    
\subfigure[Impersonation attack in face recognition task.]{\label{figure:reid1}\includegraphics[width=94mm]{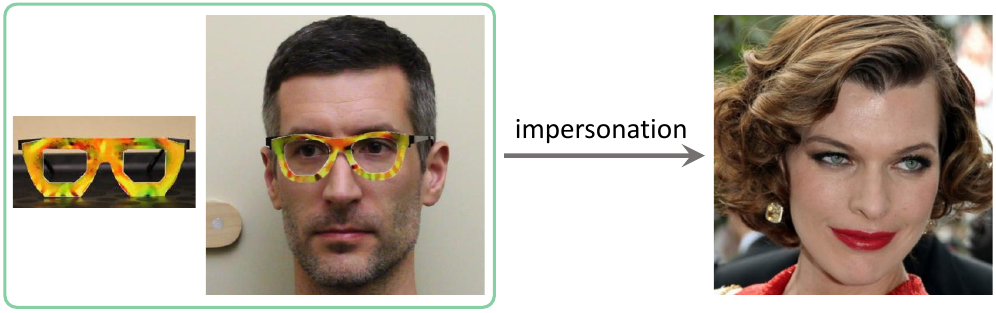}} \hspace{4mm}
\subfigure[Impersonation attack in person re-identification task.]{\label{figure:reid2}\includegraphics[width=70mm]{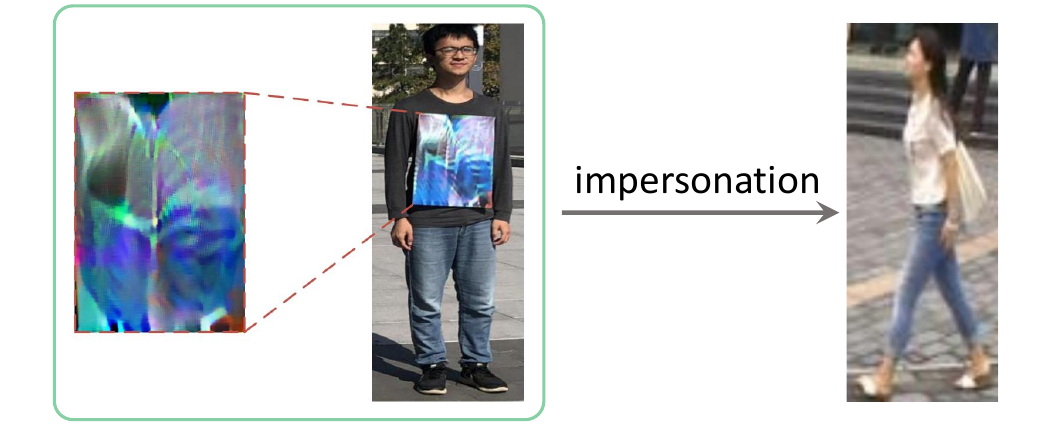}} 
\caption{Display of the physical adversarial attack in re-identification (Re-ID) tasks. Adapted from AdvEyeglass~\cite{sharif2016accessorize} (a) and AdvPattern~\cite{wang2019advpattern} (b).}
\label{figure:re-identification}
  \vspace{-0.4cm}
\end{figure*}

Lane detection is important for autonomous driving because it supports steering decisions. {\color{black}Jing \textit{et al.}~\cite{272270} pioneered the investigation into the security of lane detection modules in real vehicles, focusing on the Tesla Model S.} We term their method "Adversarial Markings" as it utilizes small markings on the road surface to mislead the vehicle's visual system. Extensive experiments show that Tesla Autopilot is vulnerable to Adversarial Markings in the physical world and follows the fake lane into oncoming traffic.

Zolfi \textit{et al.}~\cite{zolfi2021translucent} designed a universal perturbation, called TransPatch, to fool the detector for all instances of a specific object class while maintaining the detection of other objects. TransPatch is a type of colored translucent sticker, which performs attacks by attaching this special sticker to the camera lens, resulting in disturbing the camera's imaging. 
In addition, Giulio~\textit{et al.}~\cite{272218} introduced SLAP, a light-based technique enabling physical attacks in self-driving scenarios. Using a projector, SLAP projects a specific pattern onto a Stop Sign, causing YOLOv3~\cite{redmon2018yolov3} and Mask-RCNN~\cite{he2017mask} to misdetect the targeted object in moving vehicle environments. {\color{black}However, such methods may raise suspicion among human drivers. To address this, Zhu \textit{et al.}~\cite{zhu2023tpatch} proposed TPatch, innovatively designing a triggered physical adversarial patch. TPatch exhibits malicious behavior only when triggered by acoustic signals; otherwise, it behaves benignly. This approach offers a novel perspective on enhancing the stealthiness of adversarial patches.}


\subsection{Attacks on Re-Identification Tasks}
	\label{sec:reid}

In this section, we review physical adversarial attacks on re-identification (Re-ID) tasks (see Fig. 10). 
TABLE~\ref{tab:ReIDtask} presents the comparative results of these methods based on the \textit{hiPAA} metric.

\subsubsection{Face Recognition}
Face Recognition Systems (FRS) are widely used in surveillance and access control~\cite{MobileSec,NEUROTechnology}. It is valuable to explore the potential risks of FRS.
{\color{black}Sharif \textit{et al.}~\cite{sharif2016accessorize} developed a groundbreaking method to attack the face recognition algorithm by printing a pair of eyeglass frames.} 
The person who wears the adversarial eyeglasses is able to evade being recognized or impersonate another individual. 
{\color{black}The non-printability score (NPS) is defined to ensure the perturbations are printable, and it is widely used as a loss function during adversarial perturbation optimization.}
They demonstrate how an attacker that is unaware of the system’s internals is able to achieve inconspicuous impersonation under a commercial FRS~\cite{faceplusplus}.

\setlength{\tabcolsep}{4.7pt}
\begin{table}[t]\scriptsize
\centering
\caption{Comparison of the \textit{hiPPA} metric among attack methods for both the face recognition task and person Re-ID task. We highlight the minimum and maximum values using \textcolor{blue}{blue} and \textcolor{red}{red}, respectively.}
\vspace{-0.1mm}
 \begin{threeparttable}
\begin{tabular}{lp{0.4cm}p{0.4cm}p{0.4cm}p{0.4cm}p{0.4cm}p{0.4cm}c}
\toprule
 \multirow{2}{*}{Methods} & \multicolumn{6}{c}{Hexagonal Score} & \multirow{2}{*}{hiPAA}\\ \cmidrule(lr){2-7}
                    & Eff. & Rob.  &  Ste.   & Aes.  & Pra.  & Eco. & \\
  \midrule
AdvEyeglass~\cite{sharif2016accessorize} \tiny{CCS16} & 1.00 & 0.33 &0.65 &0.67 & 0.89 &0.99& 0.75\\
AdvEyeglass+ \cite{sharif2019general} \tiny{TOPS19} & 1.00 & 0.67 & 0.65 & 0.67  & 1.00 &0.99& \textcolor{red}{0.83}\\
Advhat~\cite{komkov2021advhat} \tiny{ICRP20} & 1.00 & 0.83 & 0.25 & 0.47 & 0.69 &0.99& 0.73\\
ALPA~\cite{nguyen2020adversarial}  \tiny{CVPR20} & 1.00 & 0.50 & 0.26 & 0.25 & 0.66 &0.92& 0.64\\
CLBAAttack \cite{singh2021brightness} \tiny{BIOSIG21} & 0.95 & 0.33 & 0.29 & 0.25 & 0.63 &0.99& 0.60\\
AdvMask~\cite{zolfi2021adversarial}  \tiny{EP21} & 0.96 & 1.00 & 0.23 & 0.20 & 0.86  &0.98& 0.74\\
AdvMakeup~\cite{yin2021adv} \tiny{IJCAI21} & 0.40 & 0.33 & 0.88 & 0.68 & 0.65 &0.98& 0.59\\
TAP~\cite{xiao2021improving}  \tiny{CVPR21} & 1.00 & 0.17 & 0.43 & 0.63 & 0.61 &0.99& 0.64\\
AdvSticker~\cite{wei2022adversarial}  \tiny{TPAMI22} & 0.98 & 0.67 & 0.69 & 0.88 & 0.65 &0.99& 0.82\\
SOPP~\cite{wei2022simultaneously}  \tiny{TPAMI22} & 0.96 & 0.83 & 0.42 & 0.63 & 0.65 &0.99& 0.76\\
SLAttack~\cite{Li_2023_CVPR}  \tiny{CVPR23} & 0.65 & 0.67 & 0.43 & 0.46  & 0.65 &0.29& 0.56\\
AT3D~\cite{Yang_2023_CVPR} \tiny{CVPR23} & 0.48 & 0.67 & 0.27 & 0.47 & 0.65 &0.99& \textcolor{blue}{0.54}\\
{\color{black}DOPatch~\cite{wei2023distributional} \tiny{arXiv23}} & 0.88 & 0.83 & 0.42 & 0.63 & 0.65 & 0.99 & 0.74\\
 \midrule
AdvPattern~\cite{wang2019advpattern} \tiny{ICCV19} & 0.69 & 0.50 & 0.26 & 0.25 & 0.69 &0.99& 0.55\\
\bottomrule
\end{tabular}
 \end{threeparttable}
\label{tab:ReIDtask}
  \vspace{-0.4cm}
\end{table}

Pautov \textit{et al.}~\cite{8958134} explored physical attacks on ArcFace~\cite{Deng_2019_CVPR} using adversarial patches. They designed a cosine similarity loss to minimize the similarity between the patched photo and the ground truth. The generated gray patch is easily printable. They tested the patch in three forms: eyeglasses, and stickers on the nose and forehead. Numerical experiments demonstrated effective real-world attacks on ArcFace. 
Light-based attacks have shown feasibility in classification tasks~\cite{gnanasambandam2021optical}. For FRS, Nguyen \textit{et al.}~\cite{nguyen2020adversarial} designed a real-time adversarial light projection attack using an off-the-shelf camera-projector setup, targeting state-of-the-art FRS such as FaceNet~\cite{Schroff_2015_CVPR} and SphereFace~\cite{Liu_2017_CVPR}.

\begin{figure*}
\centering    
\subfigure[Optical flow estimation]{\label{figure:ofe}\includegraphics[width=37.8mm]{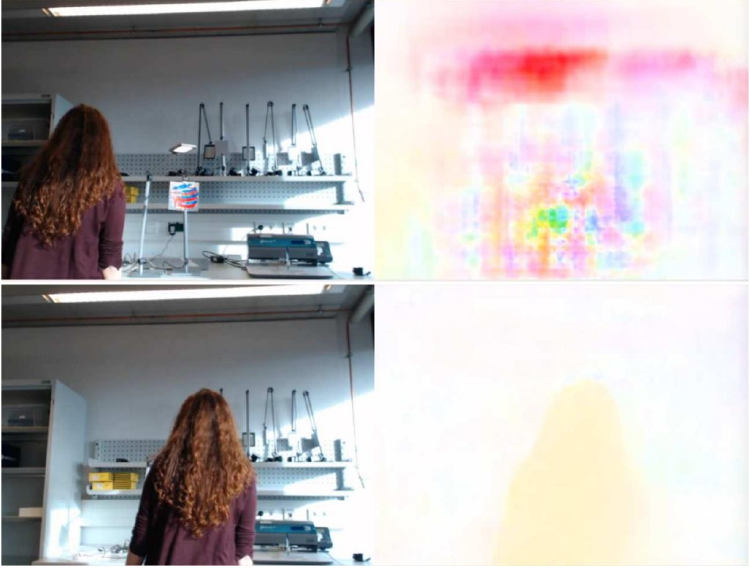}} \hspace{1mm}
\subfigure[Crowd counting]{\label{figure:crowd}\includegraphics[width=50mm]{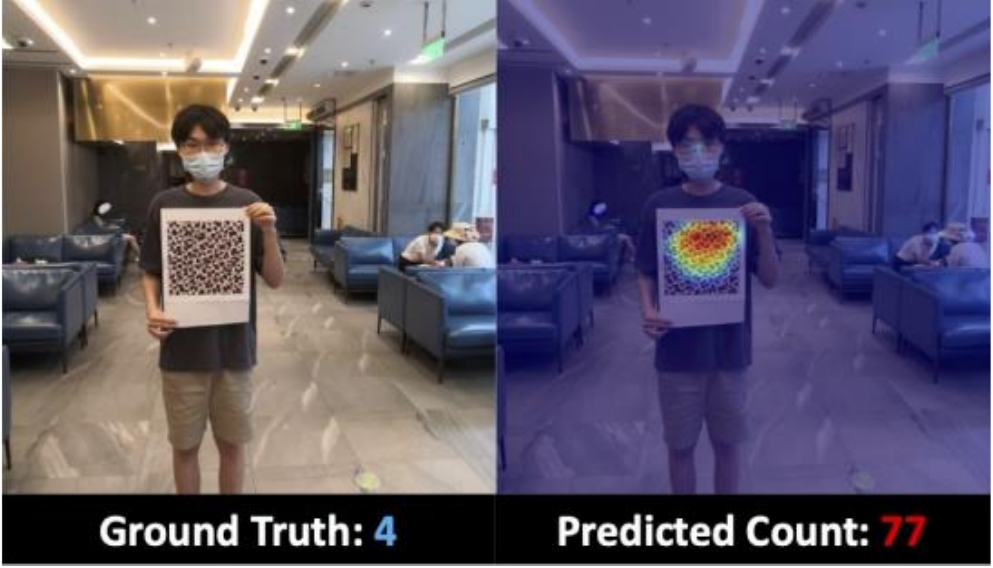}} \hspace{1mm}
\subfigure[Monocular depth estimation]{\label{figure:MDE}\includegraphics[width=51.8mm]{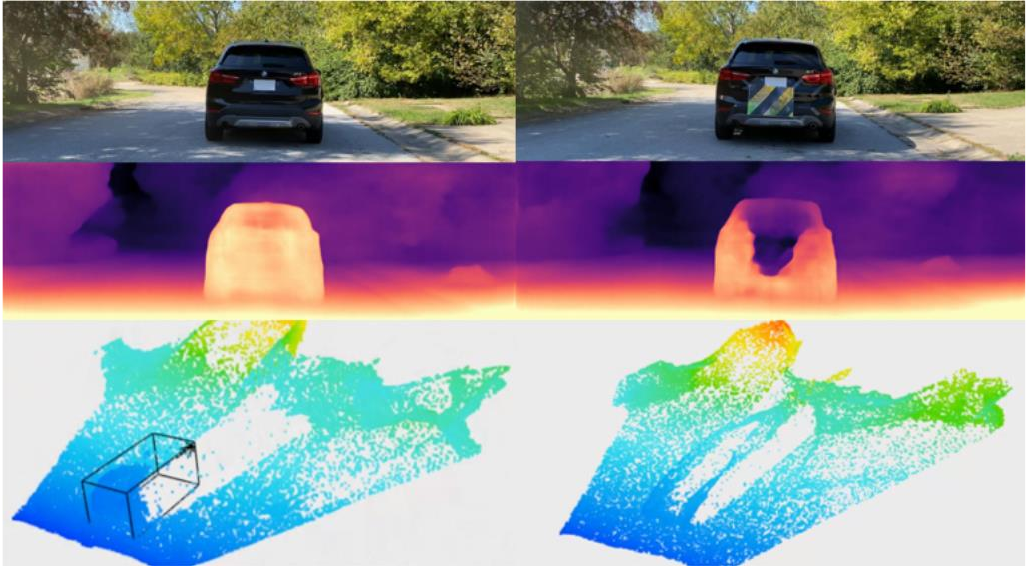}} \hspace{-1mm}
\subfigure[Semantic segmentation]{\label{figure:SSeg}\includegraphics[width=34.3mm]{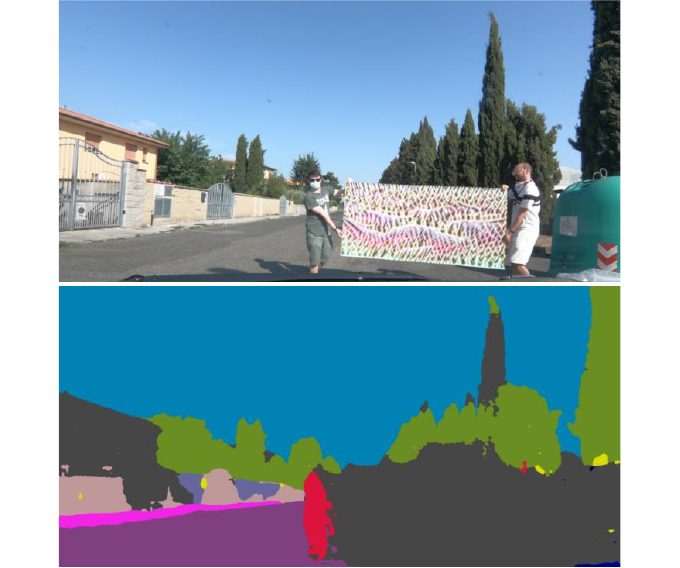}}
\caption{Display of the physical adversarial attack in other tasks. Adapted from FlowAttack~\cite{ranjan2019attacking} (a), PAP~\cite{liu2021harnessing} (b), OAP~\cite{cheng2022physical} (c), and RWAEs~\cite{nesti2022evaluating} (d).}
\label{figure:Others}
  \vspace{-0.4cm}
\end{figure*}

Advhat~\cite{komkov2021advhat} implements an easily reproducible physical adversarial attack on the state-of-the-art public Face ID system~\cite{2014Face,Deng_2019_CVPR}. In the digital space, Advhat uses Spatial Transformer Layer (STL)~\cite{NIPS2015_33ceb07b} to project the obtained sticker on the image of the face. In the physical space, Advhat launches attacks by wearing a hat with a special sticker on the forehead area, which significantly reduces the similarity to the ground truth class. 

Yin \textit{et al.}~\cite{yin2021adv} proposed AdvMakeup, a unified adversarial face generation method that addresses a common and practical scenario: applying makeup to eye regions to deceive FRS while maintaining a visually inconspicuous appearance, resembling natural makeup.
Concretely, AdvMakeup first introduces a makeup generation module, which can add natural eye shadow over the orbital region. Then, a task-driven fine-grained meta-learning adversarial attack strategy guarantees the attacking effectiveness of the generated makeup. Experimental results show that the AdvMakeup' attack effectiveness is substantially higher than Advhat~\cite{komkov2021advhat} and AdvEyeglass~\cite{sharif2016accessorize}.

\subsubsection{Person Re-Identification}
Person Re-ID is the task of identifying and tracking an individual of interest across multiple non-overlapping cameras~\cite{wang2019learning}.
This task plays an important role in surveillance and security applications.
Wang \textit{et al.}~\cite{wang2019advpattern} were the first and only ones to propose a physical attack on the Re-ID model, known as AdvPattern. They accomplished evasion and impersonation attacks by formulating distinct optimization objectives. 
As shown in Fig.~\ref{figure:reid2}, AdvPattern employs adversarial patches featuring specially crafted patterns as the adversarial medium, which are affixed to a person's chest.
The method degrades the rank-1 accuracy of person Re-ID models from 87.9\% to 27.1\% and under impersonation attack.
This easily implementable approach exposes the vulnerability of the DNNs-based Re-ID system.


{\subsection{Attacks on Other Tasks}
	\label{sec:other}}
In addition to the three aforementioned mainstream tasks, physical adversarial attacks extend to {\color{black}eight} niche tasks, encompassing
optical flow estimation~\cite{ranjan2019attacking}, steering angle prediction~\cite{kong2020physgan}, crowd counting~\cite{liu2021harnessing}, segmentation~\cite{nesti2022evaluating}, object tracking~\cite{wiyatno2019physical,ding2021towards}, 
monocular depth estimation~\cite{cheng2022physical}, image captioning~\cite{zhang2023capatch}, and {\color{black}X-ray detection}~\cite{liu2023x}.
Table~\ref{tab:nichetasks} presents the comparative results based on the \textit{hiPAA} metric.

\noindent{\textbf{Optical Flow Estimation}}
(OFE) aims to measure the pixel 2D motion of an image sequence~\cite{ilg2017flownet}. As shown in Fig.~\ref{figure:ofe}, Ranjan \textit{et al.}~\cite{ranjan2019attacking} proposed FlowAttack to perturb the OFE models. 
FlowAttack utilizes the gradients from pre-trained optical flow networks to update adversarial patches.
Experimental results show that FlowAttack can cause large errors for encoder-decoder networks but not strongly affect spatial pyramid networks. 
This phenomenon demonstrates the correlation between network structure and vulnerability.

\noindent{\textbf{Crowd Counting}}
aims to estimate the number of individuals within images or videos, with significant applications in public safety and traffic management~\cite{jiang2020attention}. Liu \textit{et al.}~\cite{liu2021harnessing} proposed a Perceptual Adversarial Patch (PAP) for attacking crowd-counting systems in the real world. 
PAP generates an adversarial patch by maximizing the model loss, leading the target victim model to overestimate the count by up to 100 on 80\% of the samples (see Fig.~\ref{figure:crowd}).

\noindent{\textbf{Monocular Depth Estimation}}
(MDE) aims to estimate the distance between the camera and a target object, which is crucial for autonomous driving~\cite{Wang_2019_CVPR}. Recently, Cheng \textit{et al.}~\cite{cheng2022physical} introduced an attack named OAP against MDE models (see Fig.\ref{figure:MDE}). OAP employs a rectangular patch region optimization method to find the optimal patch-pasting region, resulting in over 6 meters mean depth estimation error and 93\% ASR in downstream tasks. {\color{black}Despite its effectiveness, OAP relies on 2D image patches and cannot achieve multi-viewpoint attacks. To address this limitation, Zheng \textit{et al.}~\cite{zheng2024physical} proposed $3D^2$Fool, which integrates UV mapping into adversarial texture optimization, creating robust 3D camouflage textures capable of making the car vanish.}

\noindent{\textbf{Semantic Segmentation}}
aims to classify each pixel into predefined categories without distinguishing between individual object instances~\cite{he2017mask}.
Nesti \textit{et al.}~\cite{nesti2022evaluating} crafted adversarial patches to perturb the semantic segmentation models. 
As shown in Fig.~\ref{figure:SSeg}, they created a large adversarial patch, measuring 1m $\times$ 2m, which disrupts the predictions of segmentation models in the physical world.
The adversarial patch is optimized using pixel-wise cross-entropy loss on the pre-trained ICNet~\cite{zhao2018icnet}. Meanwhile, they built abundant and diverse scenes by the CARLA Simulator~\cite{dosovitskiy2017carla} for scene-specific attacks.
Experimental results show that their attack method can reduce the model's accuracy in the digital space, but in the real world, the attack is greatly downgraded. 

\noindent{\textbf{Steering Angle Prediction}} 
assists autonomous driving systems in making informed decisions~\cite{chen2015deepdriving,prakash2021multi,chitta2021neat}.
To evaluate the safety and robustness of this task, Kong \textit{et al.}~\cite{kong2020physgan} introduced PhysGAN, a method that generates physically resilient adversarial examples to deceive autonomous steering systems.
As shown in Fig.~\ref{figure:PhyGAN}, 
by utilizing the discriminator within the GAN framework to assess the visual disparities between adversarial roadside signs and their original counterparts, PhysGAN can generate realistic adversarial examples.
Meanwhile, it can maintain attack effectiveness continuously across all frames throughout the entire trajectory.

\setlength{\tabcolsep}{4.7pt}
\begin{table}[t]\scriptsize
\centering
\caption{Comparison of the hiPPA metric among attack methods for {\color{black}eight} niche tasks.}
\vspace{-0.1mm}
 \begin{threeparttable}
\begin{tabular}{lp{0.4cm}p{0.4cm}p{0.4cm}p{0.4cm}p{0.4cm}p{0.4cm}c}
\toprule
 \multirow{2}{*}{Methods} & \multicolumn{6}{c}{Hexagonal Score} & \multirow{2}{*}{hiPAA}\\ \cmidrule(lr){2-7}
                    & Eff. & Rob.  &  Ste.   & Aes.  & Pra.  & Eco. & \\
  \midrule
PAT~\cite{wiyatno2019physical} \tiny{ICCV19} & 0.60 & 0.67 & 0.27 & 0.27 & 0.87 &0.29& 0.51\\
MTD~\cite{ding2021towards}\tiny{AAAI21} & 0.74 & 0.67 & 0.27 & 0.40 & 0.69 &0.99& 0.62\\
 \midrule
OAP~\cite{cheng2022physical} \tiny{ECCV22} & 0.94 & 1.00 & 0.87 & 0.61 & 0.65 & 0.98 & 0.88\\
{\color{black}$3D^2$Fool~\cite{zheng2024physical} \tiny{CVPR24}} & 0.95 & 1.00 & 0.59 & 0.73 & 0.71 & 0.79 & 0.83\\
 \midrule
FlowAttack~\cite{ranjan2019attacking} \tiny{ICCV19} & 1.00 & 0.67 & 0.25 & 0.25 & 0.64 &0.99& 0.67\\
PhysGAN~\cite{kong2020physgan} \tiny{CVPR20} & 1.00 & 0.67 & 0.65 &0.89 &0.65 &0.95& 0.81\\
RWAEs~\cite{nesti2022evaluating} \tiny{WACV22} & 1.00 & 0.33 & 0.20 & 0.46 & 0.25 &0.95& 0.57\\
PAP~\cite{liu2021harnessing} \tiny{CCS22} & 1.00 & 0.33 & 0.64 & 0.47  & 0.61 &0.99& 0.70\\
CAPatch~\cite{zhang2023capatch} \tiny{USENIX23} & 1.00 & 1.00 & 0.20 & 0.20 & 0.63 & 0.99 & 0.72 \\ 
{\color{black}X-Adv~\cite{liu2023x} \tiny{USENIX23}} & 0.74 & 0.33 & 0.93 & 0.86 & 0.69 & 0.93 & 0.72 \\

 \bottomrule
\end{tabular}
 \end{threeparttable}
\label{tab:nichetasks}
  \vspace{-0.4cm}
\end{table}

\noindent{\textbf{Object Tracking}}
aims to detect moving objects and track them from frame to frame~\cite{yilmaz2006object}.
Wiyatno \textit{et al.}~\cite{wiyatno2019physical} proposed the first physical adversarial attack on this task.
Specifically, they perform optimization to create a distinctive pattern, which is then presented on a large monitor as a background. When a person moves in front of the monitor, the tracker tends to prioritize locking onto the background and disregards the person.
Subsequently, Ding \textit{et al.}~\cite{ding2021towards} proposed a patch-based attack method to launch universal physical attacks on single object tracking. 
As shown in Fig.~\ref{figure:MTD}, in the presence of the patch, the tracker neglects the originally tracked object.
These explorations raise security concerns for real-world visual tracking.

\noindent{\textbf{Image Captioning}}
focuses on generating a description of an image, which requires recognizing the important objects, their attributes, and their relationships in an image~\cite{hossain2019comprehensive}.
Inspired by patch-based attacks, Zhang \textit{et al.}~\cite{zhang2023capatch} designed CAPatch, a method capable of inducing errors in final captions within real-world scenarios (see Fig.~\ref{figure:CAPatch}). 
CAPatch deceives image captioning systems, causing them to produce a specified caption or conceal certain keywords.
In contrast to existing attack methods, this study represents the initial endeavor to employ an adversarial patch against multi-modal artificial intelligence systems.

{\color{black}\noindent{\textbf{X-ray Detection}}
is widely used in security screening to identify prohibited items in safety-critical scenarios~\cite{tao2021towards}. Liu \textit{et al.}~\cite{liu2023x} pioneered the exploration of physical adversarial attacks in X-ray imaging. They introduced X-Adv, a technique for creating physically plausible adversarial metal objects. When positioned near the targeted prohibited item, these objects enable the item to evade detection. X-Adv exposes vulnerabilities in X-ray detection systems, emphasizing the necessity for enhanced robustness.}

%% file: 6discussion.tex
\section{Discussion}
	\label{sec:discussion}

During the development of this paper, we have observed the diversity and broad scope of physical adversarial attacks. Despite the growth in published works in recent years, there are still gaps to be explored. Here, we discuss the current challenges and opportunities in this field.

\subsection{Current Challenges}
\subsubsection{Existing Domain Gaps}
\label{subsubsection:domain}
The workflow of physical adversarial attacks (see Fig.~\ref{figure:workflowPPA}) reveals a process where attackers first design in the digital space, deploy in the physical space, and ultimately execute attacks in the digital domain. This workflow involves the transformation between the digital and physical domains. 
{\color{black}Some researchers have recognized this; for instance, Jan \textit{et al.}~\cite{jan2019connecting} designed a D2P network to model the transformation of images from the digital domain to the physical domain. 
However, the current research has paid limited attention to addressing domain gaps, resulting in the unstable attack performance of many methods.
The present practices in physical adversarial attacks face significant challenges in reliability and reproducibility.}

\begin{figure}
\centering    
\subfigure[Steering Angle Prediction]{\label{figure:PhyGAN}\includegraphics[width=28.8mm]{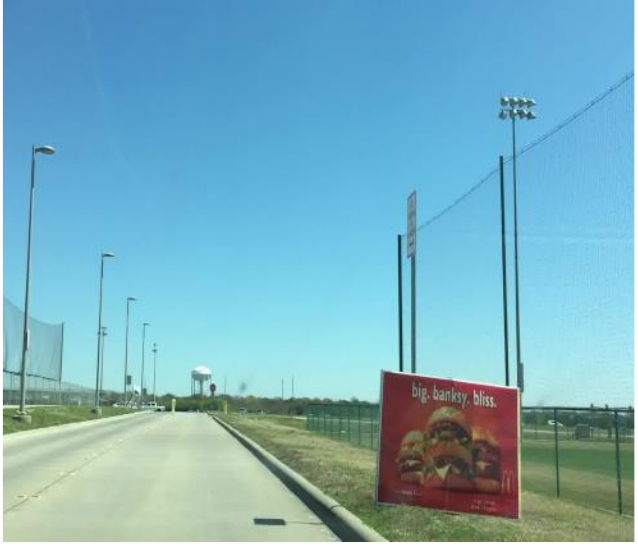}} \hspace{-0.7mm}
\subfigure[Object Tracking]{\label{figure:MTD}\includegraphics[width=28mm]{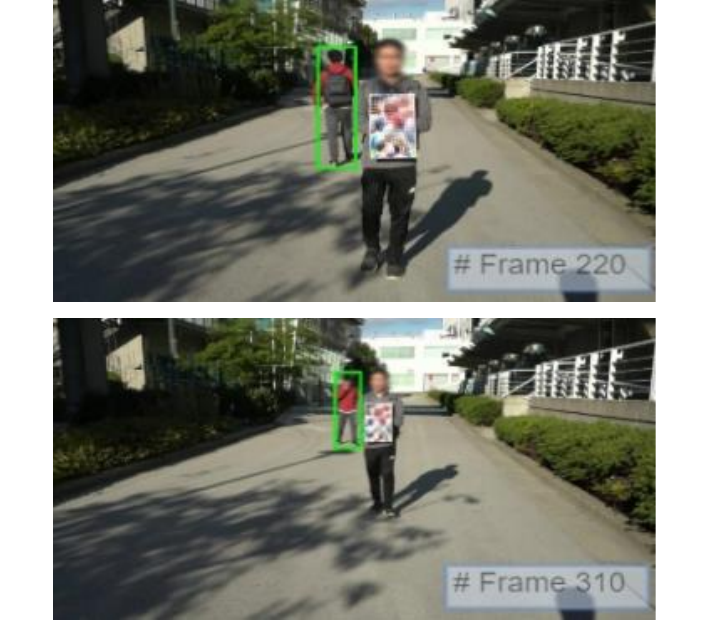}} \hspace{-0.7mm}
\subfigure[Image Captioning]{\label{figure:CAPatch}\includegraphics[width=28.1mm]{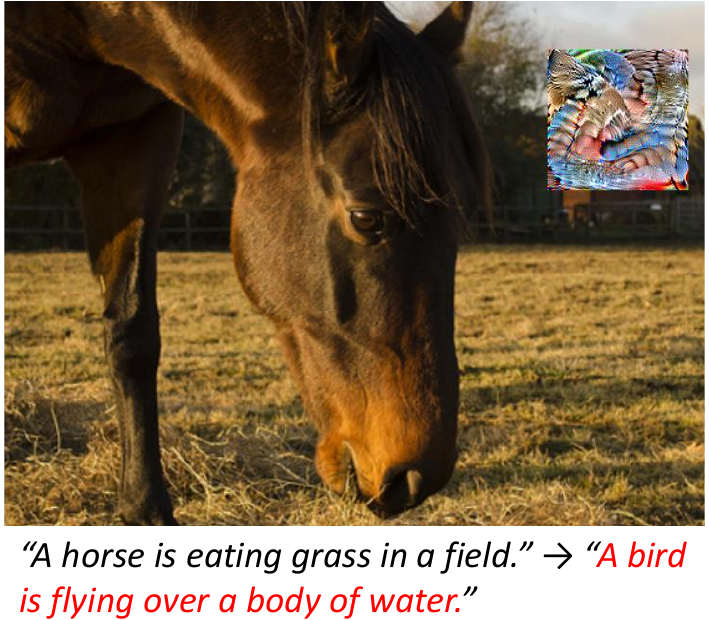}}
\caption{Display of the physical adversarial attack across three CV tasks: steering angle prediction, object tracking, and image captioning. Adapted from PhysGAN~\cite{kong2020physgan} (a), MTD~\cite{ding2021towards} (b), and CAPatch~\cite{zhang2023capatch} (c).}
\label{figure:threeothers}
  \vspace{-0.4cm}
\end{figure}

\subsubsection{Uncontrollable Evaluation Settings}
\label{subsubsection:evaluation}
Most existing works evaluate their physical adversarial attack methods in the real world using the adversarial mediums they manufacture. The real-world environment is dynamic, and the process of crafting adversarial mediums involves subjective factors, e.g., the material of the clothing, the quality of the printing, \textit{etc}, all of which are uncontrollable.
Future work with reliable and controllable evaluation setups is anticipated.

\subsection{Future Work}
\subsubsection{New Adversarial Medium}
In this survey, we define the adversarial medium as the object that carries the adversarial perturbations in the physical world. The adversarial medium plays a significant role in performing a physical attack. From the above discussion, we see that different tasks have different requirements for the adversarial medium, and the suitable adversarial medium can improve the performance of the attack in solving the trilemma, i.e., effectiveness, robustness, and stealthiness. The attacks using patches~\cite{brown2017adversarial}, light~\cite{nguyen2020adversarial}, camera ISP~\cite{phan2021adversarial}, makeup~\cite{yin2021adv}, 3D-printed object~\cite{athalye2018synthesizing}, clothing~\cite{xu2020adversarial}, \textit{etc}, emerged in turn. Recently, the Laser Beam~\cite{duan2021adversarial} and small lighting bulbs have been used to deceive the DNNs-based models, which inspire novel attack methods and expose the potential risks of DNNs-based applications.

{\color{black}\subsubsection{Cross-Domain Physical Adversarial Attacks}}
{\color{black}
To address the challenges mentioned in Sec.~\ref{subsubsection:domain}, attackers need to consider two domain gaps.

\textit{Digital-to-physical domain gap}
arises in Step 2 (see Fig.~\ref{figure:workflowPPA}), specifically denoting the procedure in which attackers manufacture physical perturbations based on digital perturbations.
A typical example is the printing loss proposed by Sharif~\textit{et al.}~\cite{sharif2016accessorize}, which specifically refers to the inability to accurately and reliably reproduce colors due to the smaller color space of printing devices compared to the RGB color space. They introduced the non-printability score (NPS) to address this issue.
Additionally, Jan~\textit{et al.}~\cite{jan2019connecting} employed an image-to-image translation network to model the digital-to-physical transformation.
Moreover, the adversarial medium, materials, and certain physical constraints all influence this domain gap. A more detailed consideration will facilitate effective cross-domain attacks.

\textit{Physical-to-digital domain gap} arises in Step 3 (see~Fig.~\ref{figure:workflowPPA}). Adversarial perturbations carried by the adversarial medium are captured by cameras in the real world, converted into digital images, and then used to attack DNNs-based models. 
Throughout this process, there exists a domain gap between the transformations from physical perturbations to digital images.
Phan \textit{et al.}~\cite{phan2021adversarial} have studied physical adversarial attacks under specific ISP conditions, but they did not explore the performance of attacks across different ISPs.
Differentiable ISP simulation or camera simulation is an ideal solution. Combining these simulators will enable the generated adversarial perturbations to maintain attack stability across various hardware imaging devices.
}

{\color{black}
\subsubsection{Reproducible Evaluation}
To address the challenges mentioned in Sec.~\ref{subsubsection:evaluation}, researchers should be required to disclose more experimental details. On one hand, this includes production details such as materials, size, and manufacturing processes of the adversarial medium. On the other hand, it involves the environmental conditions of physical experiments, such as lighting, background, and shooting distance. While these details may seem trivial and easily overlooked, they are indispensable for a comprehensive evaluation. Otani \textit{et al.}~\cite{otani2023toward} have proposed a template for researchers to reference, thereby promoting fairness and reproducibility in evaluation.
In the field of physical adversarial attacks, a template for disclosing experimental details is anticipated in the future.
}

\subsubsection{Physical World Simulation}
The key characteristic of physical adversarial attacks is their real-world feasibility. Precisely simulating the physical environment enhances attacks' robustness in dynamic settings. Simulation engines like Unreal Engine and Unity offer various conditions for attacks, such as lighting, backgrounds, camera distances, and view angles. Most existing methods use these simulators to assess attack efficacy~\cite{zhang2018camou,duan2022learning}. However, due to non-differentiability, they cannot be used in end-to-end optimization for adversarial perturbations. Beyond basic operations like rotation, noise addition, affine transformations, and occlusions~\cite{thys2019fooling,tan2021legitimate,hu2022adversarial}, integrating advanced physical scene simulation methods into the attack pipeline is crucial for considering dynamic settings during adversarial perturbation design.

{\color{black}
\subsubsection{Hexagonal Physical Adversarial Attacks}
The proposed evaluation metric, \textit{hiPAA}, evaluates physical adversarial attacks from six perspectives. However, during the evaluation, we observed that most existing methods lack a comprehensive consideration~\cite{ranjan2019attacking,zhu2022infrared}. They tend to focus on individual perspectives while neglecting others. Some approaches involve tradeoffs between individual dimensions, such as effectiveness and stealthiness~\cite{tan2021legitimate,hu2021naturalistic}. In real-world applications, physical adversarial attacks often need to excel from all six perspectives. 
Future work with holistic considerations is expected, advancing and evaluating their methods across various perspectives.
}

\subsubsection{Physical Adversarial Attacks on New Tasks}
As described in this survey, the current mainstream physical adversarial attack methods are oriented to tasks such as person detection~\cite{huang2020universal,tan2021legitimate}, traffic sign detection~\cite{220580Song,272218}, face recognition~\cite{sharif2016accessorize,8958134}, \textit{etc}. Although many fields have been covered, some tasks have not yet been explored. For example, Cheng \textit{et al.}~\cite{cheng2022physical} recently proposed an adversarial patch attack against monocular depth estimation (MDE), which is a critical vision task in real-world driving scenarios. It is the first time to propose an attack on MDE. Besides, we consider that domains with the following two characteristics can be explored for physical adversarial attacks: \textbf{1)} using the DNNs techniques, and \textbf{2)} applying in the physical world. Such as trajectory prediction~\cite{gu2021densetnt}, pose estimation~\cite{liu2022recent}, action recognition~\cite{wang2021action}, \textit{etc}.

%% file: 7conclusion.tex
{\section{Conclusion}
	\label{sec:conclusion}}

Physical adversarial attacks have cast a shadow over the reliability of deep neural networks, raising security concerns. Consequently, extensive research has proposed various methods for real-world attacks across multiple tasks.
We have provided an overview of the field of physical adversarial attacks on computer vision tasks, covering classification, detection, re-identification, and some niche tasks, with a focus on the adversarial mediums and a comprehensive evaluation. 
We first propose a general workflow for launching a physical adversarial attack, underlining the important role of the adversarial medium.
Additionally, we have devised a new metric termed \textit{hiPPA}, systematically quantifying and assessing attack methods from six distinct perspectives. Correspondingly, we present comparative results for existing methods, offering valuable insights for future improvements.
Many challenges remain ahead, and we hope that this paper can motivate further discussion in this field and provides important guidance for future research, ultimately advancing the safety and reliability of machine vision systems.

{\section{Acknowledgements}
	\label{sec:Ack}}
This work was supported by Hubei Key R\&D Project (2022BAA033), National Natural Science Foundation of China (62171325), the Fundamental Research Funds for the Central Universities, Peking University, JSPS KAKENHI (JP23K24876), and JST ASPIRE (JPMJAP2303).